\documentclass{article} % For LaTeX2e
\usepackage{iclr2027_conference,times}

\usepackage{amsmath,amsfonts,bm}

\def\eqref#1{equation~\ref{#1}}
\def\1{\bm{1}}

\DeclareMathAlphabet{\mathsfit}{\encodingdefault}{\sfdefault}{m}{sl}
\SetMathAlphabet{\mathsfit}{bold}{\encodingdefault}{\sfdefault}{bx}{n}

\usepackage{hyperref}
\usepackage{url}
\usepackage{booktabs}       % professional-quality tables
\usepackage{amsfonts}       % blackboard math symbols
\usepackage{nicefrac}       % compact symbols for 1/2, etc.
\usepackage{microtype}      % microtypography
\usepackage{xcolor}         % colors
\usepackage{graphicx}
\usepackage{amsmath,amssymb,amsthm}
\usepackage{algorithm}
\usepackage{algorithmic}
\usepackage{float}
\usepackage{bm}
\usepackage{thmtools}
\usepackage{thm-restate}

\title{Flux as a Codec: Zero-Shot Lossy Compression with Flux}

\author{Fuma Kimishima \\
Hosei University, Tokyo, Japan \\
\texttt{fuma.kimishima.3c@stu.hosei.ac.jp}
\And
Jinjia Zhou \\
Hosei University, Tokyo, Japan \\
\texttt{zhou@hosei.ac.jp}
}

\iclrfinalcopy % Show author names in the arXiv preprint.
\begin{document}

\maketitle
\lhead{Preprint}

\begin{abstract}
We propose FluxC, a zero-shot codec that uses Flux as a compression prior. Although DiffC has shown that pretrained diffusion models can serve as zero-shot codecs, we ask how the choice of coding path and reconstruction method affects rate--distortion--perception behavior.
We analyze FluxC along three axes: coding cost and bit allocation, reconstruction method, and low rate distortion. At matched signal-to-noise ratio (SNR), we show that the path choice does not affect the ideal information rate, but changes how bits are allocated across time steps. For reconstruction, we compare ancestral sampling, one-step ODE reconstruction, and many-step ODE reconstruction. Our analysis shows that ODE-based reconstruction can achieve lower distortion than ancestral sampling, especially in the low rate regime. Experiments on Gaussian sources and natural images support this prediction. Moreover, FluxC performs competitively with, and in some cases outperforms, recent neural image compression methods, suggesting that Flux can serve as a principled zero-shot codec.
\end{abstract}

\section{Introduction}
Probabilistic modeling and lossless compression are two views of the same likelihood-optimization problem. If a sender and receiver share a distribution $p$, then arithmetic coding can encode an input $\mathbf{x}$ using approximately $-\log_2 p(\mathbf{x})$ bits. Learning $p$ by maximum likelihood therefore reduces the expected ideal codelength assigned by the model. This principle underlies flow models compressed with local bits-back coding~\citep{ho2019compression}, and language models used as arithmetic-coding predictors \citep{deletang2024language,haddadi2026comparing}.

Lossy compression keeps this probabilistic-coding viewpoint but changes the object being encoded. The sender need not describe $\mathbf{x}$ itself; it instead communicates a lower-rate random variable from which the receiver reconstructs an approximation of $\mathbf{x}$. DiffC \citep{theis2022lossy,liu2025lossy} makes this connection concrete for diffusion models \citep{sohlDickstein2015deep,ho2020denoising}. It uses reverse channel coding (RCC)~\citep{li2018strong,theis2022algorithms,flamich2022fast} to communicate samples from the sender-side Gaussian kernels along the variance preserving (VP) path, with a coding cost characterized by the negative evidence lower bound (NELBO)~\citep{theis2022lossy}. Thus, the diffusion models can be viewed not only as a generative prior, but also as specifying a lossy codec.
Because lossy compression discards information about the source, reconstruction quality must be assessed beyond codelength alone. Standard criteria include distortion, which measures pointwise fidelity such as mean squared error (MSE), and perception, or realism, which measures the distance between the data distribution and the reconstructed one. It is known that these quantities are linked by the rate--distortion--perception trade-off~\citep{blau2018perception,blau2019rethinking}.

Modern image generation has increasingly moved toward Flow Matching \citep{lipman2023flow} and Rectified Flow \citep{liu2023flow}, which learn a velocity field that transports between distributions along an optimal transport (OT) path. Flux is a prominent model in this
family~\citep{flux2024}, but its existing use in DiffC reparameterizes
the Flux velocity field as a DDPM style prediction and performs coding
on a normalized VP path~\citep{liu2025lossy}, instead of the OT path~\citep{liu2025lossy} (dubbed as DiffC Flux in this paper). 
Although the DiffC family has been shown to achieve strong distortion performance, it remains unclear whether Flux can define a codec directly on its native OT path. This motivates studying whether Flux can instead define a zero-shot codec directly on its native OT path, and how this choice affects its rate--distortion--perception behavior.

We propose FluxC, a zero-shot codec that defines its Markov transitions
directly on the Flux OT path, communicates noisy variables through RCC,
and reconstructs from the communicated variables using ODE-based
decoders. To characterize its coding performance, we analyze this codec from three perspectives. First, we study coding cost and bit allocation. VP and OT paths have the same ideal information rate at matched signal-to-noise ratio (SNR), but allocate bits differently across coding steps. Second, we compare reconstruction methods, including one-step and many-step ODE reconstruction, and relate them to the DiffC Flux ancestral sampling baseline. Third, we analyze the low rate regime and show that, under the assumptions of our analysis, deterministic ODE reconstruction admits an order \(\sqrt{R}\) distortion improvement, whereas ancestral sampling improves at order \(R\), where \(R\) denotes the rate in bits.
Empirically, FluxC is competitive with existing generative compression
baselines and achieves strong distortion performance in the ultra low rate regime.

\section{Preliminaries}
\label{sec:preliminaries}

Unless otherwise stated, uppercase bold symbols such as $\mathbf{X}$ and $\mathbf{Z}_t$ denote random variables, while lowercase bold symbols such as $\mathbf{x}$ and $\mathbf{z}_t$ denote their realizations. All information quantities and coding costs are measured in bits. Therefore, logarithms in mutual information and KL divergences are base 2.
We defer a detailed discussion of related work to Appendix~\ref{app:related-work}.

\subsection{DDPMs}
\label{subsec:ddpm}

Denoising diffusion probabilistic models (DDPMs) define a Markov forward process that gradually corrupts data into Gaussian noise \citep{sohlDickstein2015deep,ho2020denoising}. We use latent variables $\mathbf{z}_t$ for $t=0,\ldots,T$, where $\mathbf{z}_0=\mathbf{x}_0$ is a sample from the data distribution and $\mathbf{z}_T$ is approximately standard Gaussian. For a variance-preserving (VP) path, let $\beta_t\in(0,1)$ be the variance increment, define $a_t=1-\beta_t$ and $\bar a_t=\prod_{s=1}^t a_s$, and set $\alpha_t^{\rm VP}=\sqrt{\bar a_t}$ and $\sigma_t^{\rm VP}=\sqrt{1-\bar a_t}$. The forward marginal is
$q^{\rm VP}(\mathbf{z}_t\mid \mathbf{x}_0)
=
\mathcal{N}\!\left(
\alpha_t^{\rm VP}\mathbf{x}_0,
(\sigma_t^{\rm VP})^2 I
\right)$, or equivalently
$\mathbf{z}_t
=
\alpha_t^{\rm VP}\mathbf{x}_0
+
\sigma_t^{\rm VP}\boldsymbol{\epsilon}$
where
$\boldsymbol{\epsilon}\sim\mathcal{N}(\mathbf{0},I)$ and $(\alpha_t^{\rm VP})^2+(\sigma_t^{\rm VP})^2=1$.
Consequently, the marginal variance scale of $\mathbf{z}_t$ remains
approximately constant as the signal coefficient decreases and the noise coefficient increases.

For the path-level comparisons below, we also use the continuous time VP-SDE parameterization. Let $B(t)=\int_0^t \beta(s)ds$. Then $(\alpha_t^{\rm VP})^2=e^{-B(t)}$, $(\sigma_t^{\rm VP})^2=1-e^{-B(t)}$, and the corresponding SNR is 

\begin{equation}
    \gamma_t^{\rm VP}=\frac{(\alpha_t^{\rm VP})^2}{(\sigma_t^{\rm VP})^2}=\frac{e^{-B(t)}}{1-e^{-B(t)}}.
    \label{eq:vp-snr}
\end{equation}

DDPMs generate samples by learning Gaussian reverse kernels that map
$\mathbf{z}_T$ back to $\mathbf{z}_0$ through the sequence
$\mathbf{z}_{T-1},\ldots,\mathbf{z}_0$. Since the marginal reverse conditional $q^{\rm VP}(\mathbf{z}_t\mid\mathbf{z}_{t+1})$ is generally intractable, DDPMs learn a Gaussian kernel $p^{\rm VP}_\theta(\mathbf{z}_t\mid\mathbf{z}_{t+1})$ using the tractable forward posterior $q^{\rm VP}(\mathbf{z}_t\mid\mathbf{z}_{t+1},\mathbf{x}_0)$. With $\alpha_{t+1|t}^{\rm VP}=\alpha_{t+1}^{\rm VP}/{\alpha_t^{\rm VP}}$ and $({\sigma_{t+1|t}^{\rm VP}})^2=({\sigma_{t+1}^{\rm VP}})^2-({\alpha_{t+1|t}^{\rm VP}})^2({\sigma_t^{\rm VP}})^2$, these kernels are

\begin{equation}
\begin{aligned}
q^{\rm VP}(\mathbf{z}_t\mid \mathbf{z}_{t+1},\mathbf{x}_0)
&=
\mathcal{N}\!\left(
\boldsymbol{\mu}_q(\mathbf{z}_{t+1},\mathbf{x}_0, \alpha_t^{\rm VP}, \sigma_t^{\rm VP};t),
\Sigma_t
\right),
\\
p^{\rm VP}_\theta(\mathbf{z}_t\mid \mathbf{z}_{t+1})
&=
\mathcal{N}\!\left(
\boldsymbol{\mu}_\theta(\mathbf{z}_{t+1}, t),
\Sigma_t
\right).
\label{eq:ddpm-reverse-kernels}
\end{aligned}
\end{equation}

where $\boldsymbol{\mu}_\theta(\mathbf{z}_{t+1},t)
=
\boldsymbol{\mu}_q(
\mathbf{z}_{t+1},
\hat{\mathbf{x}}_{\theta}^{\rm VP}(\mathbf{z}_{t+1}, \alpha_{t+1}^{\rm VP}, \sigma_{t+1}^{\rm VP}, t+1),
\alpha_t^{\rm VP},
\sigma_t^{\rm VP};
t
)$.
Under the fixed-variance parameterization of~\citep{ho2020denoising}, the reverse variance is set to the forward posterior variance $\Sigma_t$~\citep{ho2020denoising}. Thus learning the reverse kernel amounts to learning its mean. Because the forward process is linear Gaussian, the forward posterior mean admits the closed form
\begin{equation}
\boldsymbol{\mu}_q(\mathbf{z}_{t+1},\mathbf{x}_0, \alpha_t, \sigma_t;t)
=
\frac{\alpha_t{\sigma_{t+1|t}}^2}{{\sigma_{t+1}}^2}\mathbf{x}_0
+
\frac{\alpha_{t+1|t}{\sigma_t}^2}{{\sigma_{t+1}}^2}\mathbf{z}_{t+1}.
\label{eq:ddpm-forward-posterior-mean}
\end{equation}
The unavailable clean sample $\mathbf{x}_0$ is replaced by the
denoised estimate $\hat{\mathbf{x}}_{\theta}$ constructed from the predicted noise $\hat{\boldsymbol{\epsilon}}_\theta$:

\begin{equation}
\hat{\mathbf{x}}_{\theta}(\mathbf{z}_{t+1}, \alpha_{t+1}, \sigma_{t+1}, t+1)
=
\frac{
\mathbf{z}_{t+1}
-
\sigma_{t+1}\hat{\boldsymbol{\epsilon}}_\theta(\mathbf{z}_{t+1},t+1)
}{
\alpha_{t+1}
}.
\label{eq:ddpm-denoising-prediction}
\end{equation}
Thus the KL-divergence between $p^{\rm VP}_\theta(\mathbf{z}_t\mid\mathbf{z}_{t+1})$ and $q^{\rm VP}(\mathbf{z}_t\mid\mathbf{z}_{t+1},\mathbf{x}_0)$ depends on the model only through this mean~\citep{ho2020denoising}:

\begin{equation}
\begin{aligned}
&D_{\mathrm{KL}}\!\left(
    q^{\mathrm{VP}}(\mathbf{z}_t \mid \mathbf{z}_{t+1}, \mathbf{x}_0)
    \,\middle\|\,
    p^{\mathrm{VP}}_\theta(\mathbf{z}_t \mid \mathbf{z}_{t+1})
\right) \\
&\quad =
\frac{1}{2}
\Bigl\|
    \boldsymbol{\mu}_q(
        \mathbf{z}_{t+1}, \mathbf{x}_0,
        \alpha_t^{\mathrm{VP}}, \sigma_t^{\mathrm{VP}}; t
    ) -
    \boldsymbol{\mu}_\theta(
        \mathbf{z}_{t+1},
        \alpha_t^{\mathrm{VP}}, \sigma_t^{\mathrm{VP}}; t
    )
\Bigr\|_{\Sigma_t^{-1}}^2
+ \mathrm{const}.
\end{aligned}
\label{eq:ddpm-kl-mean-regression}
\end{equation}

The variational training objective for DDPMs is obtained by summing these reverse-transition KL terms, together with the endpoint Gaussian terms, over the noising chain:

\begin{equation}
\begin{aligned}
\mathcal{L}_{\rm NELBO}
&=
D_{\rm KL}\!\left(
q^{\rm VP}(\mathbf{z}_T\mid\mathbf{x}_0)
\,\|\,
p^{\rm VP}(\mathbf{z}_T)
\right) \\
&+
\sum_{t=0}^{T-1}
D_{\rm KL}\!\left(
q^{\rm VP}(\mathbf{z}_t\mid\mathbf{z}_{t+1},\mathbf{x}_0)
\,\|\,
p^{\rm VP}_\theta(\mathbf{z}_t\mid\mathbf{z}_{t+1})
\right)
-
\log p^{\rm VP}_\theta(\mathbf{x}_0\mid\mathbf{z}_1).
\end{aligned}
\label{eq:diffc-nelbo}
\end{equation}

Sampling can then be performed either by ancestral sampling from $p^{\rm VP}_\theta(\mathbf{z}_t\mid\mathbf{z}_{t+1})$ or by the corresponding probability-flow ODE~\citep{song2021score}.

\subsection{DiffC}
\label{subsec:diffc}
DiffC~\citep{theis2022lossy, liu2025lossy} interprets the variational optimization of diffusion as a coding scheme. Its communication primitive is reverse channel coding (RCC)~\citep{li2018strong, theis2022algorithms, flamich2022fast}. If the sender samples $\mathbf{z}_t\sim q^{\rm VP}(\mathbf{z}_t\mid\mathbf{z}_{t+1},\mathbf{x}_0)$ and the receiver shares a proposal $p^{\rm VP}_\theta(\mathbf{z}_t\mid\mathbf{z}_{t+1})$, $\mathbf{z}_t$ can be communicated at the cost approximately $D_{\rm KL}(q^{\rm VP}(\mathbf{z}_t\mid\mathbf{z}_{t+1},\mathbf{x}_0)\|p^{\rm VP}_\theta(\mathbf{z}_t\mid\mathbf{z}_{t+1}))$~\citep{theis2022algorithms}. 
If the sender and receiver share the Gaussian randomness, apply RCC to the intermediate samples over the \(T-1\) reverse steps, and entropy-coding the final data sample with \(p^{\rm VP}_\theta(\mathbf{x}_0\mid\mathbf{z}_0)\), then the resulting total lossless coding cost coincides with the negative ELBO in Eq.~\ref{eq:diffc-nelbo}~\citep{theis2022lossy}. Thus, training the diffusion model by maximizing ELBO is equivalent to minimizing the total coding cost induced by DiffC.
For lossy compression, DiffC reconstructs from the intermediate sample $\mathbf{x}_t$ by ancestral sampling (DiffC-A) or probability-flow ODE (DiffC-F).

\subsection{Rectified Flow}
\label{subsec:rectified-flow}

Rectified Flow learns a velocity field that transports samples between two distributions~\citep{lipman2023flow,liu2023flow}. We use the OT path parameterization with data at $t=0$ and Gaussian noise $\boldsymbol{\epsilon}\sim\mathcal{N}(\mathbf{0},I)$ at $t=1$. Let $\mathbf{X}\in \mathbb{R}^d$ denote a data sample. Then the corresponding random path is $\mathbf{Z}^{\rm OT}_t=(1-t)\mathbf{X}+t\boldsymbol{\epsilon}$.
Its target velocity, which is constant along the linear OT path, is
$\dot{\mathbf{Z}}^{\rm OT}_t:=\frac{\mathrm{d}}{\mathrm{d}t}\mathbf{Z}^{\rm OT}_t=\boldsymbol{\epsilon}-\mathbf{X}$.
A Rectified Flow model learns a velocity field
$\mathbf{v}_\theta(\mathbf{Z}_t,t)$ by minimizing
$\mathbb{E}\!\left[
\left\|\mathbf{v}_\theta(\mathbf{Z}_t,t)-\dot{\mathbf{Z}}_t\right\|_2^2
\right]$. Although this parameterization differs from DDPM noise prediction in Eq.~\ref{eq:ddpm-denoising-prediction}, the velocity can be converted into denoising and noise predictions~\citet{lai2025principles} as
\begin{equation}
\hat{\mathbf{x}}_\theta^{\rm OT}(\mathbf{z},t)
=
\mathbf{z}-t\mathbf{v}_\theta(\mathbf{z},t).
\label{eq:rf-denoising-reparam-x}
\end{equation}
\begin{equation}
\hat{\boldsymbol{\epsilon}}_\theta^{\rm OT}(\mathbf{z},t)
=
\mathbf{z}+(1-t)\mathbf{v}_\theta(\mathbf{z},t).
\label{eq:rf-denoising-reparam-eps}
\end{equation}
% This reparameterization is the bridge between Rectified Flow velocity models and the denoising quantities used below.

This reparameterization is the bridge between Rectified Flow velocity models and the denoising quantities used below.

\begin{figure*}[t]
\begin{minipage}[t]{0.49\textwidth}
\begin{algorithm}[H]
 \caption{FluxC Sender} \label{alg:ot-path-sender}
 \small
 \let\AND\relax
 \begin{algorithmic}[1]
    \STATE Send $\mathbf{z}_T\sim q_T^{\rm OT}(\cdot\mid\mathbf{x}_0)$ using $p_\theta({\mathbf{z}_T})$ \hfill{\small // shared seed}
    \FOR{$t=T-1,T-2,\ldots,1$}
       \STATE Send $\mathbf{z}_t\sim q^{\rm OT}(\mathbf{z}_t\mid\mathbf{z}_{t+1},\mathbf{x}_0)$ using $p_\theta^{\rm OT}(\mathbf{z}_t\mid\mathbf{z}_{t+1})$
    \ENDFOR
    \STATE Send $\mathbf z_0\sim
q^{\rm OT}(\mathbf z_0\mid\mathbf z_1,\mathbf x_0)$
using $p_\theta^{\rm OT}(\mathbf z_0\mid\mathbf z_1)$
    % \STATE \hfill{\small // no operation}
    % \STATE \hfill{\small // no operation}
 \end{algorithmic}
\end{algorithm}
\end{minipage}
\hfill
\begin{minipage}[t]{0.49\textwidth}
\begin{algorithm}[H]
 \caption{FluxC Receiver} \label{alg:ot-path-receiver}
 \small
 \let\AND\relax
 \begin{algorithmic}[1]
    \STATE Receive $\mathbf{z}_T$ using $p_\theta({\mathbf{z}_T})$ \hfill{\small // shared seed}
    \FOR{$t=T-1,T-2,\ldots,1$}
       \STATE Receive $\mathbf{z}_t$ using $p_\theta^{\rm OT}(\mathbf{z}_t\mid\mathbf{z}_{t+1})$
       \STATE $\hat{\mathbf{x}}=\Psi^{\rm 1}(\mathbf{z}_t, t)$ \hfill{\small // one-step lossy decode}
       \STATE $\hat{\mathbf{x}}=\Psi^{\rm ODE}(\mathbf{z}_t, t)$ \hfill{\small // many-step lossy decode}
       \ENDFOR
    \STATE Receive $\mathbf{x}_0$ with $p_\theta^{\rm OT}(\mathbf{x}_0\mid\mathbf{z}_{1})$ \hfill{\small // lossless decode}
 \end{algorithmic}
\end{algorithm}
\end{minipage}
\vspace{-1em}
\end{figure*}

\section{FluxC: Flux as a Codec and Its Analysis}
\label{sec:transport-codec}

% We introduce Flux as a codec, dubbed as FluxC, and analyze which design choices determine its rate--distortion--perception behavior. FluxC keeps the RCC communication protocol but defines the coding state and Markov transition directly on the OT path. The rest of the section follows three axes: the coding cost and bit allocation induced by the path, the reconstruction method used by the receiver, and the low-rate behavior of these reconstruction methods.

\subsection{FluxC}
\label{subsec:ot-path-diffc}
FluxC is a zero-shot codec that performs coding along the OT path of Flux. As in DiffC~\citep{theis2022lossy}, the sender and receiver share the Gaussian random seed and use RCC to communicate intermediate noisy variables. However, using a Flux backbone does not uniquely determine the Markov reverse transition $q(\cdot\mid\mathbf{z}_{t+1},\mathbf{x}_0)$ or require the VP diffusion path. DiffC Flux~\citep{liu2025lossy} reparametrizes the Flux OT path into the VP path, and the Flux velocity is converted into a denoising prediction through Eq.~\ref{eq:rf-denoising-reparam-x}.
In contrast, FluxC keeps the Markov process on the OT path as
\begin{equation}
    q^{\rm OT}(\mathbf{z}_t\mid \mathbf{z}_{t+1}, \mathbf{x}_0) = \mathcal{N}\!\left(
\boldsymbol{\mu}_q(\mathbf{z}_{t+1},\mathbf{x}_0, \alpha_t^{\rm OT}, \sigma_t^{\rm OT};t),
\Sigma_t
\right).
\end{equation}
Note that the scheduling parameters used to compute $\boldsymbol{\mu}_q$ are $\alpha_t^{\rm OT}=1-t$ and $\sigma_t^{\rm OT}=t$. The receiver approximates this transition by
\begin{equation}
p_\theta^{\rm OT}(\mathbf{z}_t\mid\mathbf{z}_{t+1})
=
\mathcal{N}\!\left(
\boldsymbol{\mu}_{\theta}
\!\left(
\mathbf{z}_{t+1},
\hat{\mathbf{x}}_\theta^{\rm OT}(\mathbf{z}_{t+1}, t+1);
t
\right),
\Sigma_t
\right).
\label{eq:ot-receiver-transition}
\end{equation}
where \(\boldsymbol{\mu}_q\) and \(\Sigma_t\) are the Gaussian posterior mean and covariance obtained by applying the DDPM posterior formulas in Eqs.~\ref{eq:ddpm-forward-posterior-mean} and~\ref{eq:ddpm-denoising-prediction} with the OT coefficients. Equation~\ref{eq:ot-receiver-transition} uses the velocity--denoising reparameterization in Eq.~\ref{eq:rf-denoising-reparam-x}.

At each coding step, RCC communicates the sender-side sample using the receiver-side distribution, giving the ideal step cost
$
D_{\rm KL}\!\left(
q^{\rm OT}(\mathbf{z}_t\mid \mathbf{z}_{t+1},\mathbf{x}_0)
\,\|\,
p_\theta^{\rm OT}(\mathbf{z}_t\mid \mathbf{z}_{t+1})
\right).
$
Repeating this procedure over the \(T-1\) intermediate steps, together with the shared Gaussian seed and the final entropy coding term, yields the FluxC lossless coding cost, which coincides with the OT path NELBO. Algorithms~\ref{alg:ot-path-sender} and~\ref{alg:ot-path-receiver} summarize the sender and receiver in detail.
%The notation ``send/receive \(\mathbf{z}_t\) using \(q\)/\(p\)'' denotes the RCC communication protocol; see~\citep{li2018strong, theis2022algorithms, flamich2022fast} for details of RCC.

\subsection{Coding Cost and Bit Allocation}
\label{subsec:coding-cost-bit-allocation}
In this section, we ask whether the path selection affects the coding cost. We compare VP and OT through the scalar Gaussian observation
\begin{equation}
\mathbf{Z}_t=\alpha_t\mathbf{X}+\sigma_t\boldsymbol{\epsilon},
\qquad
\gamma_t=\frac{\alpha_t^2}{\sigma_t^2},
\qquad
\boldsymbol{\epsilon}\sim\mathcal{N}(\mathbf{0},I),
\label{eq:general-snr-path}
\end{equation}
where $\gamma_t$ is the SNR at $t$. Here $\alpha_t$ and $\sigma_t$ denote the coefficients of the path being compared. VP uses the DDPM coefficients $(\alpha_t^{\rm VP},\sigma_t^{\rm VP})$, whereas OT uses $(\alpha_t^{\rm OT},\sigma_t^{\rm OT})=(1-t,t)$. 
For lossless coding, both FluxC and DiffC yield coding costs that coincide with their NELBO objectives. For lossy compression, however, the rate depends on where the coding process is stopped. Since the VP and OT paths map the same time \(t\) to different noise levels, we compare their lossy coding costs at matched SNR, where their Gaussian observation channels are equivalent, and ask which components of the cost remain path dependent.

\begin{restatable}[Lossy coding cost decomposition]{observation}{lossyrate}
Let $\mathbf{X}\sim P_X$ be an $\mathbb{R}^d$ valued source random
variable satisfying $\mathbb{E}\|\mathbf{X}\|_2^2 < \infty$, $\gamma \geq 0$
be the SNR, and
$\boldsymbol{\epsilon}\sim\mathcal{N}(\mathbf{0},\mathbf{I}_d)$ be
independent Gaussian noise. We define the Gaussian observation at SNR
$\gamma$ as
$\mathbf{Y}_\gamma = \sqrt{\gamma}\,\mathbf{X}+\boldsymbol{\epsilon}$,
and its posterior mean estimator as
$\hat{\mathbf{X}}_{\mathrm{MMSE}}(\mathbf{Y}_\gamma,\gamma) := \mathbb{E}\!\left[\mathbf{X}\mid\mathbf{Y}_\gamma\right]$.
Let $\mathcal{C}_\theta(t)$ denote the lossy coding cost stopping at $t$, whose SNR is $\gamma_t$.
In the continuous SNR limit, this cost decomposes as
\begin{equation}
\mathcal{C}_\theta(t)
=
I(\mathbf{X};\mathbf{Y}_{\gamma_t})
+
\frac{1}{2\ln 2}
\int_0^{\gamma_t}
\mathbb{E}\!\left[
\left\|
\hat{\mathbf{X}}_\theta(\mathbf{Y}_\gamma,\gamma)
-
\hat{\mathbf{X}}_{\mathrm{MMSE}}(\mathbf{Y}_\gamma,\gamma)
\right\|_2^2
\right]
\mathrm{d}\gamma .
\label{eq:coding-cost-mismatch}
\end{equation}
\end{restatable}

See Appendix~\ref{app:matched-snr-coding-cost} for details.
The first term is the mutual information required to communicate the
observation at SNR $\gamma_t$ and therefore depends only on the matched SNR. The second term is the excess coding cost induced by denoiser mismatch. Thus, at matched SNR, differences in coding cost reflect how accurately the respective backbones approximate the posterior mean along the encoding path.

The model mismatch in Eq.~\ref{eq:coding-cost-mismatch} is characterized as a function of SNR.
We now specialize this term to time. Changing variables from the continuous SNR coordinate \(\gamma\) to the path time \(t\) yields the following corollary.

\begin{restatable}[FluxC coding cost as weighted velocity error]
{corollary}{fluxcmismatch}
\label{obs:fluxc-velocity-error}
For the OT path, let $t\in[0,1]$ denote the path time. The noisy state at time $t$ is
$\mathbf{Z}_t=(1-t)\mathbf{X}+t\boldsymbol{\epsilon}$, and the target velocity is
$\dot{\mathbf{Z}}_t=\boldsymbol{\epsilon}-\mathbf{X}$.
The coding cost stopping at time $t$ is
\begin{equation}
\mathcal{C}_\theta(t)
=
\frac{1}{\ln 2}
\int_t^{1}
\frac{1-s}{s}
\mathbb{E}\!\left[
\left\|
\mathbf{v}_\theta(\mathbf{Z}_s,s)-\dot{\mathbf{Z}}_s
\right\|_2^2
\right]
\mathrm{d}s .
\label{eq:fluxc-coding-cost}
\end{equation}
\end{restatable}

See Appendix~\ref{app:fluxc-weighted-velocity-cost} for details.
This corollary show that FluxC coding cost can be interpreted as evaluating the velocity regression error of Flux under the coding induced weight ${1-s}/{(s\ln 2)}$.

Although VP and OT have the same ideal coding cost (i.e. zero model mismatch) at matched SNR, they parameterize SNR differently over path time. We next examine how the information cost is distributed along each path.

\begin{figure}[t]
\centering
\includegraphics[width=0.98\linewidth]{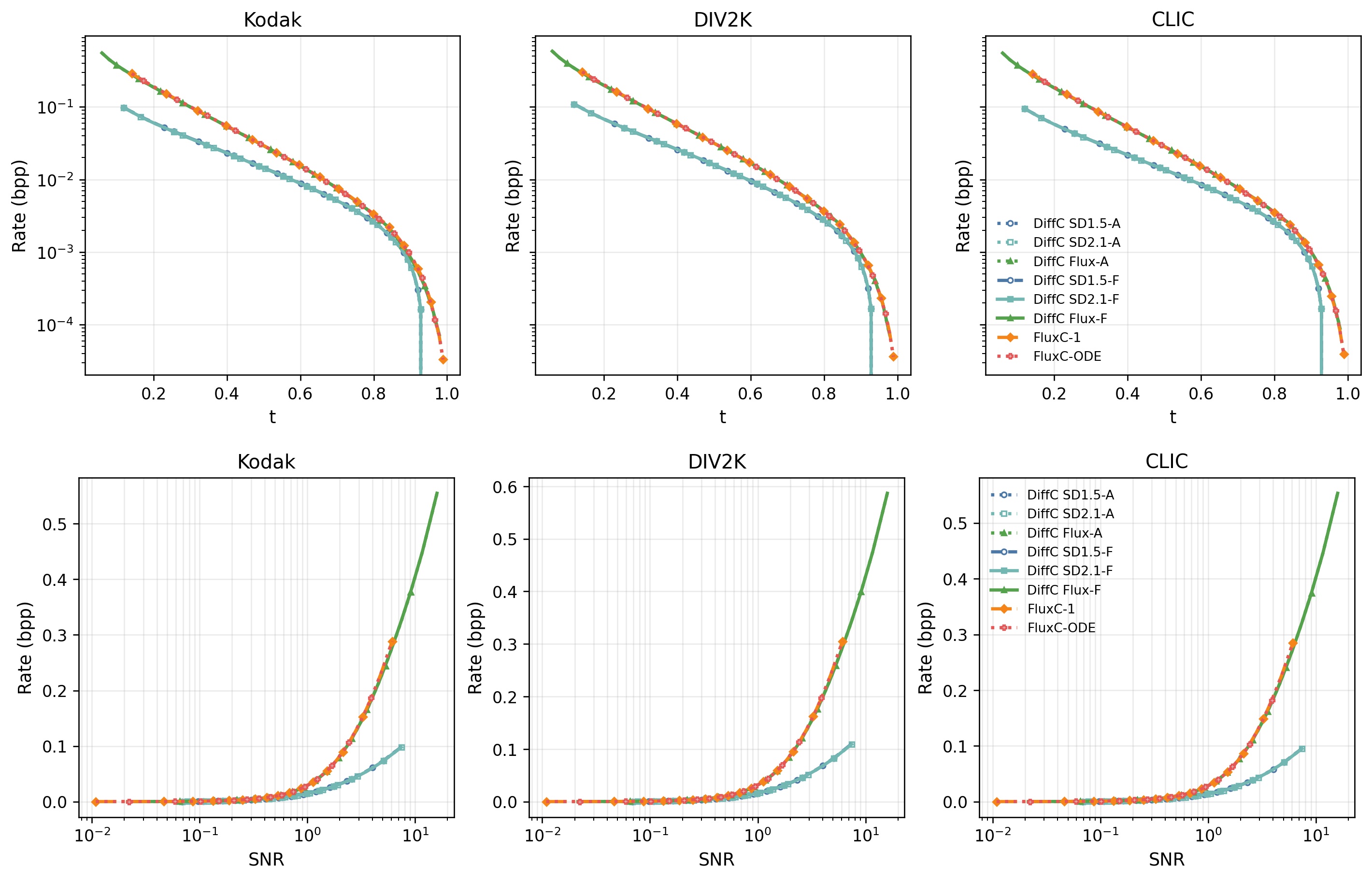}
\caption{(Top) Rate as a function of the path time \(t\) on three datasets. 
(Bottom) Rate as a function of SNR on a logarithmic scale. }
\label{fig:rate-vs-time-snr}
\end{figure}

\begin{restatable}[Bit allocation over path time]
{observation}{bitallocation}
\label{obs:path-time-bit-allocation}
For a path with coefficients $\alpha_t$ and $\sigma_t$, define its SNR as
$\gamma_t=\frac{\alpha_t^2}{\sigma_t^2}$.
The information increment per unit path time satisfies
\begin{equation}
\left|
\frac{\mathrm{d}}{\mathrm{d}t}
I(\mathbf{X};\mathbf{Y}_{\gamma_t})
\right|
=
\frac{1}{2\ln 2}
\operatorname{mmse}(\gamma_t)
\left|
\frac{\mathrm{d}\gamma_t}{\mathrm{d}t}
\right|.
\label{eq:path-parameter-bit-allocation}
\end{equation}

where $\operatorname{mmse}(\gamma)
=
\mathbb{E}\!\left[
\left\|
\mathbf{X}
-
\hat{\mathbf{X}}_{\mathrm{MMSE}}(\mathbf{Y}_\gamma,\gamma)
\right\|_2^2
\right]$.
For the continuous time VP path, let
$B(t)=\int_0^t\beta(s)\,\mathrm{d}s$. The derivatives of the VP and
OT SNRs are, respectively,
\begin{align}
\left|
\frac{\mathrm{d}\gamma_t^{\mathrm{VP}}}{\mathrm{d}t}
\right|
&=
\frac{\beta(t)e^{-B(t)}}{\left(1-e^{-B(t)}\right)^2},
\label{eq:vp-sde-snr-derivative}
\\
\left|
\frac{\mathrm{d}\gamma_t^{\mathrm{OT}}}{\mathrm{d}t}
\right|
&=
\frac{2(1-t)}{t^3}.
\label{eq:ot-snr-derivative}
\end{align}
\end{restatable}

See Appendix~\ref{app:bit-allocation-derivation} for derivation.
Although the MMSE is fixed at a given SNR, the path dependent SNR derivative $\lvert \mathrm{d}\gamma_t/\mathrm{d}t\rvert$ causes VP and OT to allocate information differently over path time.

\subsection{Reconstruction and Rate--Distortion-Perception Tradeoffs}
\label{subsec:reconstruction-methods}

We next study how the reconstruction method changes distortion and perception. We consider three reconstruction methods. First, DiffC Flux~\cite{liu2025lossy} which reparameterizes the Flux velocity field as a DDPM style denoising prediction using Eq.~\ref{eq:rf-denoising-reparam-x}, computes the reverse mean as Eq.~\ref{eq:ddpm-forward-posterior-mean}, and then samples by ancestral sampling in our settings. Second, \emph{FluxC-1} applies a one-step ODE in FluxC from the communicated time $t$ to the data endpoint. Third, \emph{FluxC-ODE} integrates the many-step ODE.
Under exact population regression, the learned field equals the
conditional mean of the target velocity,
$\mathbf{v}_t(\mathbf{z})
=\mathbb{E}[\boldsymbol{\epsilon}-\mathbf{X}
\mid\mathbf{Z}_t=\mathbf{z}]$
~\citep{liu2023flow,lipman2023flow}. Then one backward Euler step from \(t\) to \(0\) gives
\begin{equation}
\hat{\mathbf{X}}_{\mathrm{FluxC\text{-}1},\gamma_t}
:=
\mathbf{Z}_t-t\mathbf{v}_t(\mathbf{Z}_t)
=
\mathbb{E}[\mathbf{X}\mid\mathbf{Z}_t].
\label{eq:ot-one-step-posterior-mean}
\end{equation}
Thus FluxC-1 is a posterior-mean estimator (see Appendix~\ref{app:gaussian-fluxc-one-step} for derivation). Since conditional expectation is the Bayes estimator under squared-error loss~\citep{cover2006elements}, this decoder is distortion-oriented. However, a posterior mean is not generally a sample from the data distribution conditioned on the transmitted variable. It can therefore reduce pointwise distortion while shifting the reconstruction distribution, which worsens perception. 
%Appendix~\ref{app:gaussian-closed-form} makes this tradeoff explicit for Gaussian sources. 
In contrast, DiffC Flux uses ancestral sampling, while FluxC-ODE follows a deterministic transport trajectory to the data endpoint. These reconstruction methods provide the main receiver-side axis for the rate--distortion--perception comparisons.

\begin{figure}[t]
\centering
\includegraphics[width=0.98\linewidth]{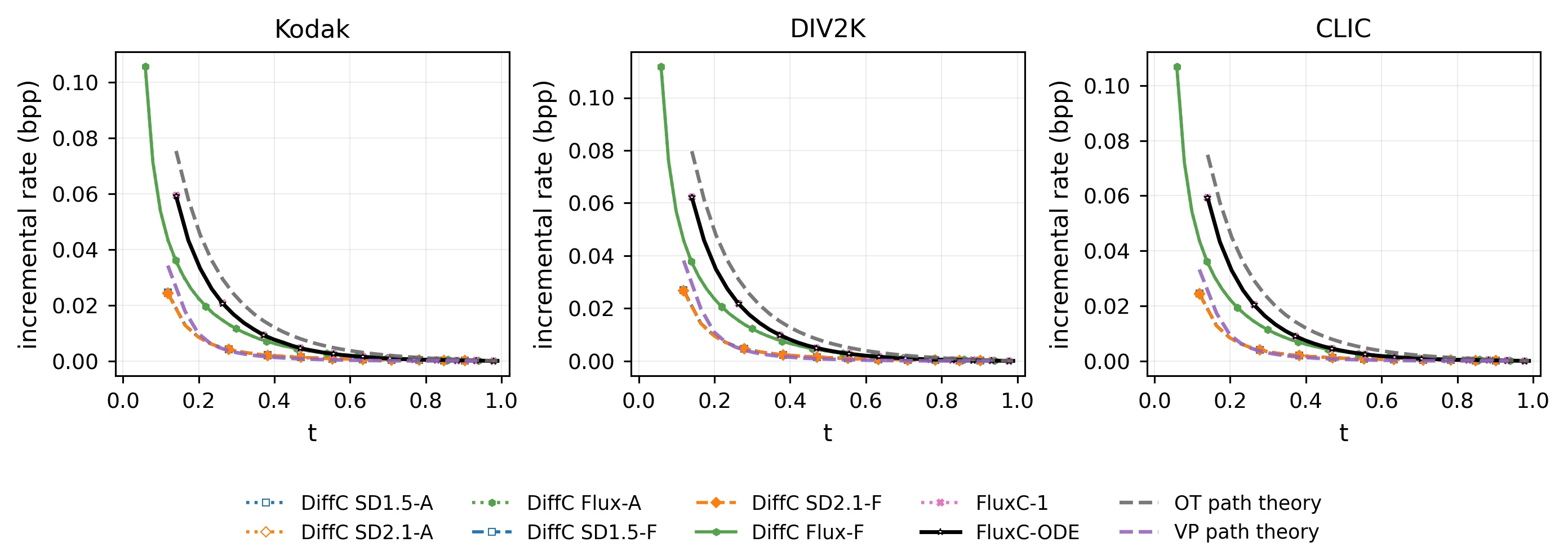}
\caption{Incremental rate as a function of \(t\). The incremental rate denotes the additional coding cost assigned to each coding step, i.e., the increase in rate between adjacent time points.}
\label{fig:incremental-rate-t}
\end{figure}

\subsection{Low Rate Analysis}
\label{subsec:reconstruction-low-rate}

Low rate compression is practically important because it reduces communication and storage cost. However, a small coding cost is useful only if the reconstruction preserves meaningful source content. This has made low bitrate distortion and perception a central focus of recent generative compression methods~\citep{careil2024perco, liu2025lossy}. We analyze this regime with finite second moment under the Gaussian observation channel induced by the communicated noisy variable.

\begin{restatable}[Low rate distortion scaling]
{result}{lowratedistortion}
\label{res:low-rate-distortion}
Let $R\geq0$ denote the rate in bits. Assume $\mathbb{E}[\mathbf{X}]=\mathbf{0}$, $0<\mathbb{E}[\|\mathbf{X}\|_2^2]<\infty$, and that the MMSE curve is right differentiable at $\gamma=0$. Let $D_0:=\mathbb{E}[\|\mathbf{X}-\widehat{\mathbf{X}}_0\|_2^2]=2\mathbb{E}[\|\mathbf{X}\|_2^2]$, where $\widehat{\mathbf{X}}_0\sim P_X$ is the reconstructed sample at $\gamma=0$ and independent of $\mathbf{X}$. Let $D_{\mathrm{ODE}}(R)$ and $D_{\mathrm{A}}(R)$ denote the MSE distortions of ideal deterministic ODE reconstruction and ancestral sampling at rate $R$, respectively. If the ideal ODE decoder preserves the source marginal, admits a one-sided $L^2$ expansion at the noise endpoint $t=1$, and has a positive first-order source response, then, as $R\downarrow0$,
\begin{equation}
D_{\mathrm{ODE}}(R)
=
D_0-c_{\mathrm{ODE}}\sqrt{R}+o(\sqrt{R}),
\qquad c_{\mathrm{ODE}}>0.
\label{eq:low-rate-ode-main}
\end{equation}
Ideal ancestral sampling satisfies
\begin{equation}
D_{\mathrm{A}}(R)
=
D_0-c_{\mathrm{A}}R+o(R),
\qquad c_{\mathrm{A}}\geq0.
\label{eq:low-rate-ancestral-main}
\end{equation}
\end{restatable}
Appendix~\ref{app:low-rate-expansion} gives the proof.
The result indicates that, near zero rate, ODE-based reconstruction can reduce distortion at order \(\sqrt R\), whereas ancestral sampling reduces only linearly in \(R\). 
% Intuitively, the ODE decoder can use even a weak observation to move the reconstruction toward the source. By contrast, ancestral sampling remains close to drawing an independent posterior sample at very low SNR, so its distortion improves only more gradually.

\begin{figure}
\centering
\includegraphics[width=0.98\linewidth]{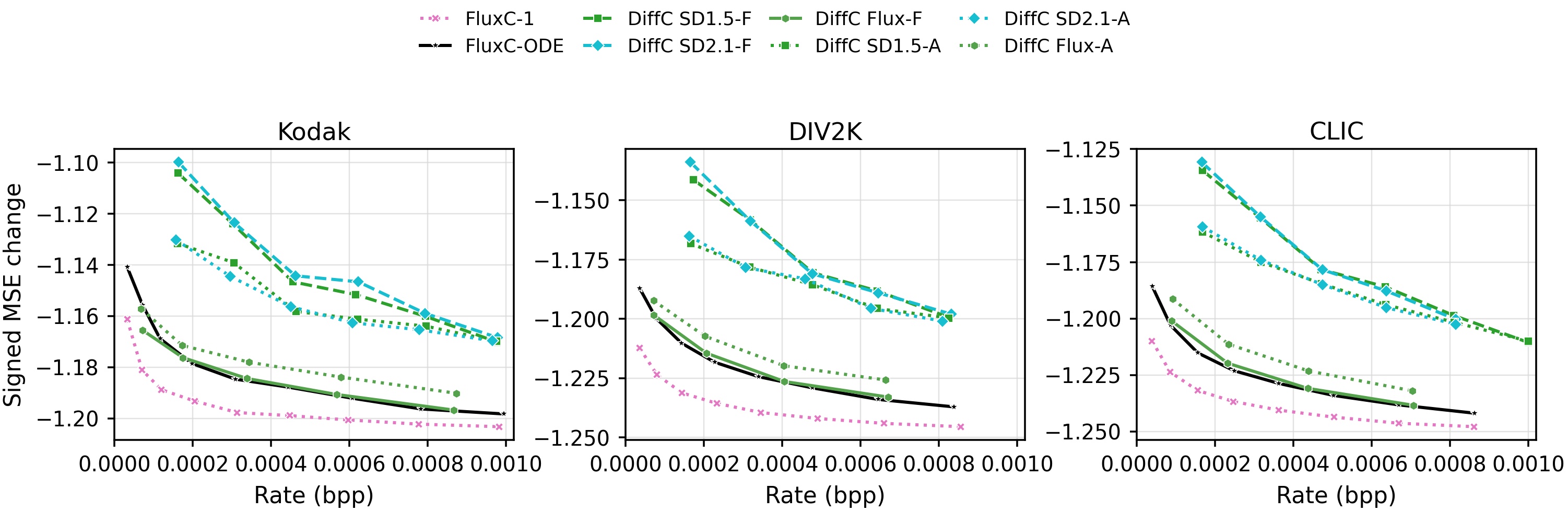}
\caption{Signed MSE change relative to the corresponding zero rate reconstruction in the ultra low rate regime (\(R\leq10^{-3}\) bpp). More negative values indicate larger distortion reductions.}
\label{fig:ultralow-distortion-improvement}
\end{figure}

\begin{figure}[t]
\centering
\includegraphics[width=0.98\linewidth]{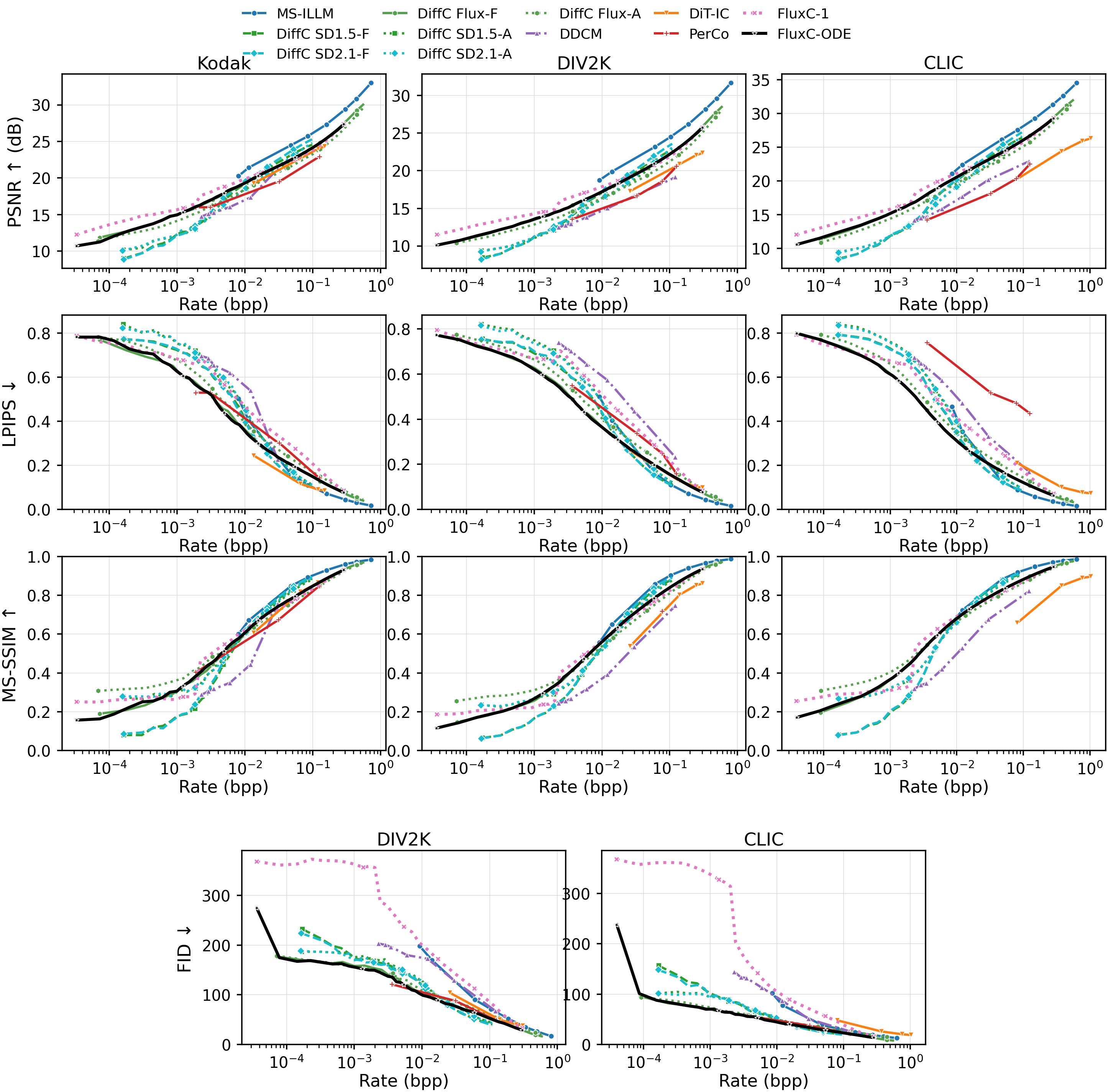}
\caption{Rate--distortion and rate--perception curves on natural images. We report PSNR, LPIPS, and MS-SSIM as distortion and FID as perception on a logarithmic rate axis.}
\label{fig:natural-rd}
\end{figure}

\section{Experimental Results}
\label{sec:experiments}
We focus on results on natural images in the main text, and Gaussian source experiments are provided in Appendix~\ref{app:experiments-gaussian}.
Because the rate--distortion and rate--perception behavior can be characterized in closed form for Gaussian sources, this setting provides a controlled test of the consistency of our theoretical analysis. The resulting experiments show that ODE based reconstruction achieves a larger distortion reduction than stochastic sampling, particularly in the low rate regime.

\paragraph{Setup.}
Due to computational constraints, we implement FluxC using the public FLUX.1-dev model~\citep{flux2024} on NVIDIA RTX 3090 GPUs with 24\,GB of memory. We follow the practical RCC implementation of the pretrained DiffC baseline~\citep{liu2025lossy}, and use 50 encoding steps and 50 decoding steps.
For the OT path, however, the SNR diverges near data endpoint \(t=0\). We therefore stop encoding before reaching the data endpoint, at approximately \(t=0.2\). For a fair comparison, we apply the same stopping rule to the DiffC variants. See details in Table~\ref{tab:encode_decode_time} in Appendix~\ref{app:computational-efficiency}.
DiffC with Stable Diffusion (SD) 1.5 or 2.1 communicates VP-path variables, DiffC Flux uses a Flux backbone in VP-normalized DiffC, FluxC-1 and FluxC-ODE are our OT-path variants and reconstruct with one-step or many-step ODE. We compare against representative neural compression baselines, including PerCo~\citep{careil2024perco}, MS-ILLM~\citep{muckley2023improving}, DDCM~\citep{ohayon2025ddcm}, and DiT-IC~\citep{shi2026ditic}.
We evaluate on Kodak~\citep{kodak1999lossless}, DIV2K~\citep{agustsson2017ntire}, and CLIC~\citep{clic2024}. To make the inputs comparable across methods, each image is center cropped to $512\times512$ before compression. We report bits per pixel (bpp) as the rate. For distortion, we use PSNR and MS-SSIM~\citep{wang2003multiscale}, where higher is better, and LPIPS~\citep{zhang2018lpips}, where lower is better. For perception, we report FID~\citep{heusel2017fid} where lower is better. Because FID is unreliable with very small sample counts, we report it only on DIV2K and CLIC.
For FID evaluation, each input and reconstruction is split into four non-overlapping $256\times256$ patches, and 10,000 patches are randomly sampled from the resulting patch pool.

\label{subsec:experiments-natural}

\begin{figure}[t]
\centering
\includegraphics[width=0.98\linewidth]{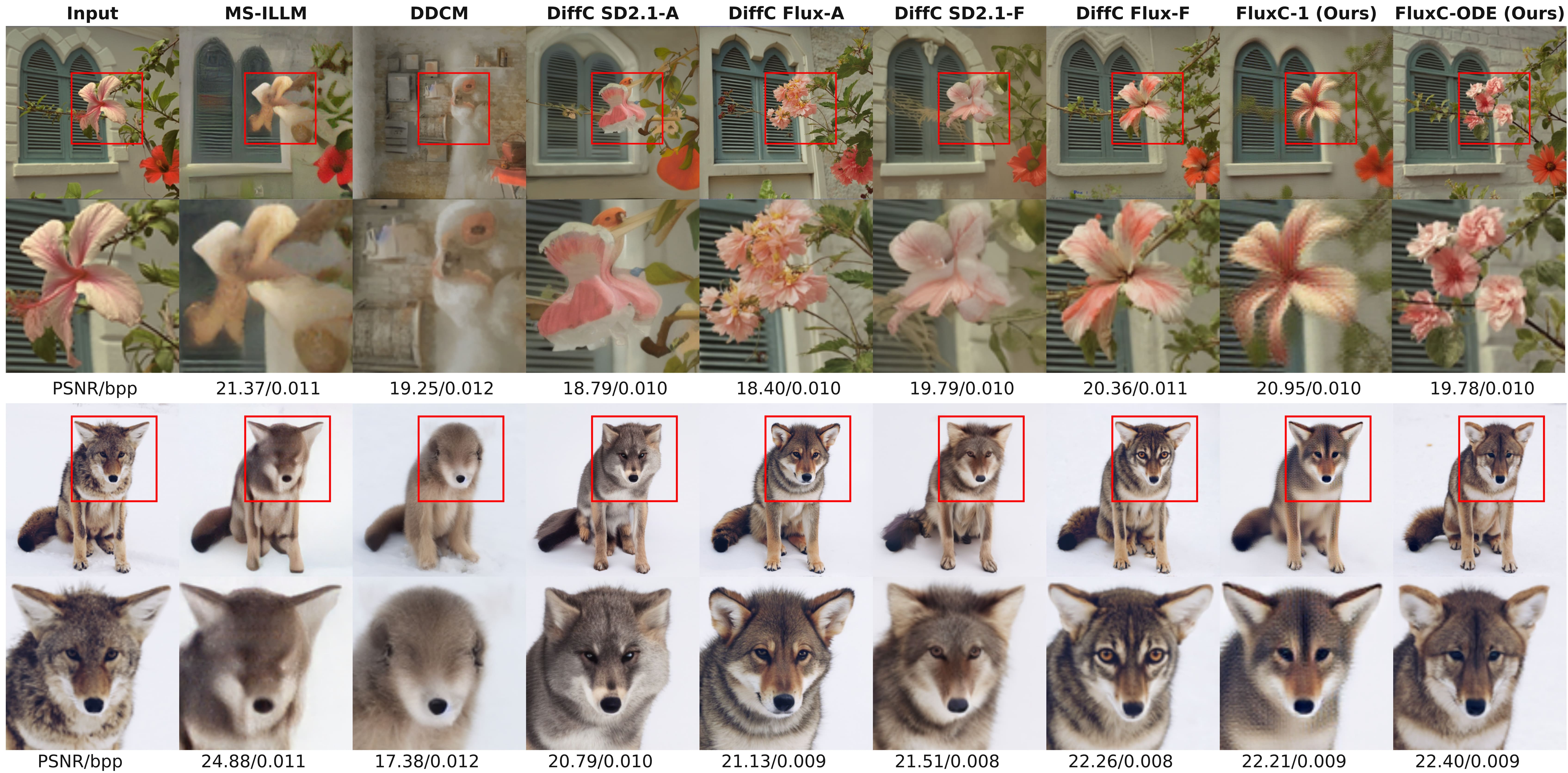}
\caption{Qualitative reconstructions at approximately $0.01$ bpp. The examples compare MS-ILLM~\citep{muckley2023improving}, DDCM~\citep{ohayon2025ddcm}, DiffC and FluxC variants}
\label{fig:kodak-reconstructions}
\end{figure}

\paragraph{Coding Cost and Bit Allocation}

\label{subsec:experiments-coding-cost-bit-allocation}

In the rate--time plot of Figure~\ref{fig:rate-vs-time-snr},
SD 1.5 and SD 2.1 produce nearly identical curves, suggesting that the allocation pattern is governed primarily by the VP path rather than the backbone. DiffC allocates most of its rate near the data endpoint, whereas FluxC distributes it over a broader range of path times.
In contrast, the rate--SNR curves of DiffC and FluxC closely agree at low SNR, consistent with Observation~\ref{obs:path-time-bit-allocation}. At high SNR, empirical coding overhead and the divergence of $\gamma_t^{\rm OT}=(1-t)^2/t^2$ as $t\to0$ lead to larger deviations from the ideal matched-SNR prediction.

The incremental--rate plot reported in Figure~\ref{fig:incremental-rate-t} shows that both DiffC and FluxC closely follow their corresponding theoretical allocation curves (Eq.~\ref{eq:vp-sde-snr-derivative} and Eq.~\ref{eq:ot-snr-derivative}, respectively), especially near the noise endpoint $t=1$. FluxC begins to allocate noticeably more rate after approximately \(t=0.6\), whereas DiffC shows a sharper increase only after approximately \(t=0.2\). 
%This indicates that the OT and VP paths distribute information over coding time in different ways. FluxC allocates information earlier along the trajectory, while DiffC concentrates more of its rate closer to the data endpoint.
We note that DiffC Flux does not exactly follow the OT path theoretical curve. Its implementation uses the diffusion time schedule and imposes a diffusion-style normalization on the OT path, namely \((\alpha_t^{\rm OT})^2+(\sigma_t^{\rm OT})^2=1\). 

\paragraph{Distortion improvement in the ultra low rate regime.}
Figure~\ref{fig:ultralow-distortion-improvement} shows the distortion
improvement in the ultra low rate regime, $R\leq10^{-3}$ bpp. For a
reconstruction $\hat{\mathbf{X}}_R$ at rate $R$, we define
$\Delta_{\mathrm{MSE}}(R):=\mathbb{E}\|\mathbf{X}-\hat{\mathbf{X}}_R\|_2^2-\mathbb{E}\|\mathbf{X}-\hat{\mathbf{X}}_0\|_2^2$, where $\hat{\mathbf{X}}_0$ is the corresponding zero-rate reconstruction; hence, more negative values indicate larger distortion reductions.
Result~\ref{res:low-rate-distortion} predicts a leading
$\sqrt{R}$ improvement for deterministic ODE reconstruction and a
linear in $R$ improvement for ancestral sampling. Consistent with this
prediction, the ODE-based methods exhibit a pronounced initial decrease
below approximately $2\times10^{-4}$ bpp, whereas the ancestral DiffC
variants decrease more gradually over the same range.

\paragraph{Rate--distortion--perception behavior.}
Figure~\ref{fig:natural-rd} summarizes the natural image rate--distortion and rate--perception behavior. 
At low rates, FluxC-1 and FluxC-ODE consistently achieve higher PSNR than the ancestral baselines, qualitatively aligning with
Result~\ref{res:low-rate-distortion}. However, DiffC-A outperforms the deterministic DiffC-F in PSNR, and the ordering also varies across metrics: ODE reconstruction performs better in LPIPS, whereas ancestral sampling performs better in MS-SSIM. Thus, the asymptotic MSE result does not imply a universal ranking for natural images or practical distortion metrics. The Gaussian-source analysis and experiments in Appendix~\ref{app:experiments-gaussian}, which satisfy
the assumptions of the result, provide a controlled verification of the predicted low rate behavior.
This distortion improvement is accompanied by worse FID than
FluxC-ODE. The smoother FluxC-1 reconstructions illustrate the expected perception cost of posterior mean decoding.
Beyond the ultra low bitrate regime, FluxC remains competitive with existing neural codecs, particularly in LPIPS and FID across a broad range of rates. 
%These results suggest that the Flux backbone is a useful prior for generative compression, although the current implementation uses the public FLUX.1-dev model without compression-specific adaptation. We expect stronger Flux-family backbones such as Flux.2~\citep{flux-2-2025} and Flux.3~\citep{blackforestlabs2026flux3} or rate-adapted reconstruction objectives to further improve the high-rate LPIPS and FID behavior.

Figure~\ref{fig:kodak-reconstructions} shows qualitative reconstructions around $0.01$ bpp. At this rate, several baseline methods lose fine details and local structure, and the generative decoders compensate by hallucinating plausible image content from their priors. FluxC-1 instead preserves more of the input-aligned structure, but outputs smoother reconstructions. FluxC-ODE produces more natural looking reconstructions, which is consistent with their respective distortion- and perception-oriented behavior.

\section{Conclusion}

We proposed Flux as a codec, FluxC, that defines the Markov transition directly on the OT path, and analyzed the consequences of this design. The analysis separates three factors. First, VP and OT variables have the same ideal information rate at matched SNR, but allocate bits differently across coding steps. Second, DiffC Flux provides a VP-normalized ancestral sampling baseline, while FluxC-1 and FluxC-ODE use OT path with deterministic ODE reconstruction. Third, the low rate expansion predicts a distortion advantage for ODE reconstruction over ancestral sampling.
Experimental results support this view: FluxC improves distortion, especially in the ultra low bit rate regime.  
Future work could investigate stronger Flux backbones, such as
Flux.2~\citep{flux-2-2025} and Flux.3~\citep{blackforestlabs2026flux3}. %Another direction is to extend the present analysis beyond Rectified Flow to other few-step generative models, including MeanFlow~\citep{geng2025mean} and flow-map models~\citep{sabour2026align}. 
%It should also clarify the rate--distortion--perception trade-off between pixel-space and latent-space codecs, since many recent image generators operate in latent spaces rather than directly in pixel space.

\subsection*{AI use statement}
In this work, generative AI tools were used to assist with implementation, to check the assumptions and intermediate derivations underlying Result~\ref{res:low-rate-distortion} and to refine the presentation of its proof.
The statement, proof strategy, and mathematical claims were
formulated by the authors, who independently verified all AI-assisted derivations.
Generative AI tools were also used to improve the clarity, correctness of expression, and readability of the manuscript. The initial draft, title, and overall structure of the paper were created by the authors. Therefore the AI tools were not used to determine the paper's scientific claims or organization.
All AI-assisted code was reviewed and tested by the authors using unit tests and the experiments reported in the paper. 

\subsection*{Ethics statement}
Although FluxC may reduce the storage and communication costs of high-resolution images, its generative decoder may hallucinate or alter visual content, making reconstructed images unsuitable for applications that require faithful preservation of evidence without additional validation.

\subsection*{Reproducibility statement}
Section~\ref{subsec:ot-path-diffc} describes the proposed algorithm and the experimental setup in Section~\ref{subsec:experiments-natural} specifies the datasets, preprocessing procedures and computational resources used in our experiments. 
Tables~\ref{tab:encode_decode_time} and~\ref{tab:fluxc_step_timing}
report the encoding and decoding times, as well as the number of
encoding steps used in our experiments.

\bibliography{iclr2027_conference}

@article{theis2022lossy,
  title = {Lossy Compression with Gaussian Diffusion},
  author = {Theis, Lucas and Salimans, Tim and Hoffman, Matthew D. and Mentzer, Fabian},
  journal = {arXiv preprint arXiv:2206.08889},
  year = {2022},
  url = {https://arxiv.org/abs/2206.08889}
}

@article{yang2024progressive,
  title = {Progressive Compression with Universally Quantized Diffusion Models},
  author = {Yang, Yibo and Will, Justus C. and Mandt, Stephan},
  journal = {arXiv preprint arXiv:2412.10935},
  year = {2024},
  url = {https://arxiv.org/abs/2412.10935}
}

@inproceedings{liu2023flow,
  title = {Flow Straight and Fast: Learning to Generate and Transfer Data with Rectified Flow},
  author = {Liu, Xingchao and Gong, Chengyue and Liu, Qiang},
  booktitle = {International Conference on Learning Representations},
  year = {2023},
  url = {https://openreview.net/forum?id=XVjTT1nw5z}
}

@inproceedings{lipman2023flow,
  title = {Flow Matching for Generative Modeling},
  author = {Lipman, Yaron and Chen, Ricky T. Q. and Ben-Hamu, Heli and Nickel, Maximilian and Le, Matt},
  booktitle = {International Conference on Learning Representations},
  year = {2023},
  url = {https://openreview.net/forum?id=PqvMRDCJT9t}
}

@misc{flux2024,
  author = {{Black Forest Labs}},
  title = {{FLUX}},
  year = {2024},
  howpublished = {\url{https://github.com/black-forest-labs/flux}}
}

@article{guo2005mutual,
  title = {Mutual Information and Minimum Mean-Square Error in Gaussian Channels},
  author = {Guo, Dongning and Shamai, Shlomo and Verdu, Sergio},
  journal = {IEEE Transactions on Information Theory},
  volume = {51},
  number = {4},
  pages = {1261--1282},
  year = {2005},
  publisher = {IEEE}
}

@misc{kimishima2026fewstep,
  title = {Few-step Generative Models as Lossy Compression},
  author = {Kimishima, Fuma and Zhou, Jinjia},
  year = {2026},
  eprint = {2606.10450},
  archivePrefix = {arXiv},
  primaryClass = {cs.CV},
  url = {https://arxiv.org/abs/2606.10450}
}

@inproceedings{sohlDickstein2015deep,
  title = {Deep Unsupervised Learning using Nonequilibrium Thermodynamics},
  author = {Sohl-Dickstein, Jascha and Weiss, Eric and Maheswaranathan, Niru and Ganguli, Surya},
  booktitle = {Proceedings of the 32nd International Conference on Machine Learning},
  pages = {2256--2265},
  year = {2015},
  volume = {37},
  series = {Proceedings of Machine Learning Research},
  publisher = {PMLR},
  url = {https://proceedings.mlr.press/v37/sohl-dickstein15.html}
}

@inproceedings{ho2020denoising,
  title = {Denoising Diffusion Probabilistic Models},
  author = {Ho, Jonathan and Jain, Ajay and Abbeel, Pieter},
  booktitle = {Advances in Neural Information Processing Systems},
  volume = {33},
  pages = {6840--6851},
  year = {2020},
  url = {https://proceedings.neurips.cc/paper/2020/hash/4c5bcfec8584af0d967f1ab10179ca4b-Abstract.html}
}

@inproceedings{deletang2024language,
  title = {Language Modeling Is Compression},
  author = {Deletang, Gregoire and Ruoss, Anian and Duquenne, Paul-Ambroise and Catt, Elliot and Genewein, Tim and Mattern, Christopher and Grau-Moya, Jordi and Wenliang, Li Kevin and Aitchison, Matthew and Orseau, Laurent and Hutter, Marcus and Veness, Joel},
  booktitle = {International Conference on Learning Representations},
  year = {2024},
  url = {https://arxiv.org/abs/2309.10668}
}

@inproceedings{haddadi2026comparing,
  title = {Comparing Text Compression Capabilities of Large Language Models with Traditional Compression Algorithms},
  author = {Haddadi, Mehran and Teahan, William John},
  booktitle = {Proceedings of the 19th Conference of the European Chapter of the Association for Computational Linguistics (Volume 4: Student Research Workshop)},
  pages = {219--232},
  year = {2026},
  publisher = {Association for Computational Linguistics},
  url = {https://aclanthology.org/2026.eacl-srw.16/}
}

@inproceedings{ho2019compression,
  title = {Compression with Flows via Local Bits-Back Coding},
  author = {Ho, Jonathan and Lohn, Evan and Abbeel, Pieter},
  booktitle = {Advances in Neural Information Processing Systems},
  year = {2019},
  url = {https://openreview.net/forum?id=BJeMOErgIH}
}

@inproceedings{song2021score,
  title = {Score-Based Generative Modeling through Stochastic Differential Equations},
  author = {Song, Yang and Sohl-Dickstein, Jascha and Kingma, Diederik P. and Kumar, Abhishek and Ermon, Stefano and Poole, Ben},
  booktitle = {International Conference on Learning Representations},
  year = {2021},
  url = {https://openreview.net/forum?id=PxTIG12RRHS}
}

@article{li2018strong,
  title = {Strong Functional Representation Lemma and Applications to Coding Theorems},
  author = {Li, Cheuk Ting and El Gamal, Abbas},
  journal = {IEEE Transactions on Information Theory},
  volume = {64},
  number = {11},
  pages = {6967--6978},
  year = {2018},
  publisher = {IEEE}
}

@inproceedings{theis2022algorithms,
  title = {Algorithms for the Communication of Samples},
  author = {Theis, Lucas and Ahmed, Noureldin},
  booktitle = {International Conference on Machine Learning},
  pages = {21308--21328},
  year = {2022},
  publisher = {PMLR},
  url = {https://proceedings.mlr.press/v162/theis22a.html}
}

@inproceedings{flamich2022fast,
  title = {Fast Relative Entropy Coding with A* Coding},
  author = {Flamich, Gergely and Markou, Stratis and Hernandez-Lobato, Jose Miguel},
  booktitle = {International Conference on Machine Learning},
  pages = {6548--6577},
  year = {2022},
  publisher = {PMLR},
  url = {https://proceedings.mlr.press/v162/flamich22a.html}
}

@misc{lai2025principles,
  title = {The Principles of Diffusion Models},
  author = {Lai, Chieh-Hsin and Chen, De-An and Chen, Han-Yi and Huang, Chin-Wei and Lin, Yen-Chen and Welling, Max and Yoon, Jong Chul and Hsieh, Cho-Jui},
  year = {2025},
  eprint = {2510.21890},
  archivePrefix = {arXiv},
  primaryClass = {cs.LG},
  url = {https://arxiv.org/abs/2510.21890}
}

@inproceedings{liu2025lossy,
  title={Lossy compression with pretrained diffusion models},
  author={Liu, Feng and others},
  booktitle={International Conference on Learning Representations},
  volume={2025},
  pages={687--702},
  year={2025}
}

@inproceedings{blau2018perception,
  title={The perception-distortion tradeoff},
  author={Blau, Yochai and Michaeli, Tomer},
  booktitle={Proceedings of the IEEE conference on computer vision and pattern recognition},
  pages={6228--6237},
  year={2018}
}

@inproceedings{blau2019rethinking,
  title={Rethinking lossy compression: The rate-distortion-perception tradeoff},
  author={Blau, Yochai and Michaeli, Tomer},
  booktitle={International Conference on Machine Learning},
  pages={675--685},
  year={2019},
  organization={PMLR}
}

@inproceedings{careil2024perco,
  title = {Towards Image Compression with Perfect Realism at Ultra-Low Bitrates},
  author = {Careil, Marlene and Muckley, Matthew J. and Verbeek, Jakob and Lathuili{\`e}re, St{\'e}phane},
  booktitle = {International Conference on Learning Representations},
  year = {2024},
  url = {https://openreview.net/forum?id=ktdETU9JBg}
}

@article{ohayon2025ddcm,
  title = {Compressed Image Generation with Denoising Diffusion Codebook Models},
  author = {Ohayon, Guy and Manor, Hila and Michaeli, Tomer and Elad, Michael},
  journal = {arXiv preprint arXiv:2502.01189},
  year = {2025},
  url = {https://arxiv.org/abs/2502.01189}
}

@article{shi2026ditic,
  title = {{DiT-IC}: Aligned Diffusion Transformer for Efficient Image Compression},
  author = {Shi, Junqi and Lu, Ming and Li, Xingchen and Ke, Anle and Zhang, Ruiqi and Ma, Zhan},
  journal = {arXiv preprint arXiv:2603.13162},
  year = {2026},
  url = {https://arxiv.org/abs/2603.13162}
}

@inproceedings{muckley2023improving,
  title={Improving statistical fidelity for neural image compression with implicit local likelihood models},
  author={Muckley, Matthew J and El-Nouby, Alaaeldin and Ullrich, Karen and Jegou, Herve and Verbeek, Jakob},
  booktitle={International Conference on Machine Learning},
  pages={25426--25443},
  year={2023},
  organization={PMLR}
}

@inproceedings{wang2024language,
  title={Language Models as Zero-shot Lossless Gradient Compressors: Towards General Neural Parameter Prior Models},
  author={Wang, Hui-Po and Fritz, Mario},
  booktitle={Advances in Neural Information Processing Systems},
  year={2024}
}

@article{mao2025lossless,
  title={Lossless Compression of Large Language Model-Generated Text via Next-Token Prediction},
  author={Mao, Yu and Pirk, Holger and Xue, Chun Jason},
  journal={arXiv preprint arXiv:2505.06297},
  year={2025}
}

@article{rinberg2026haiku,
  title={Haiku to Opus in Just 10 bits: {LLMs} Unlock Large Compression Gains},
  author={Rinberg, Roy and Carrell, Annabelle Michael and Henniger, Simon and Carlini, Nicholas and Warr, Keri},
  journal={arXiv preprint arXiv:2604.02343},
  year={2026}
}

@inproceedings{kingma2021variational,
  title = {Variational Diffusion Models},
  author = {Kingma, Diederik P. and Salimans, Tim and Poole, Ben and Ho, Jonathan},
  booktitle = {Advances in Neural Information Processing Systems},
  volume = {34},
  pages = {21696--21707},
  year = {2021},
  url = {https://openreview.net/forum?id=2LdBqxc1Yv}
}

@book{cover2006elements,
  title={Elements of Information Theory},
  author={Cover, Thomas M. and Thomas, Joy A.},
  edition={2},
  publisher={Wiley-Interscience},
  year={2006}
}

@article{kingma2013auto,
  title={Auto-encoding variational bayes},
  author={Kingma, Diederik P and Welling, Max},
  journal={arXiv preprint arXiv:1312.6114},
  year={2013}
}

@article{balle2016end,
  title={End-to-end optimized image compression},
  author={Ball{\'e}, Johannes and Laparra, Valero and Simoncelli, Eero P},
  journal={arXiv preprint arXiv:1611.01704},
  year={2016}
}

@article{balle2018variational,
  title={Variational image compression with a scale hyperprior},
  author={Ball{\'e}, Johannes and Minnen, David and Singh, Saurabh and Hwang, Sung Jin and Johnston, Nick},
  journal={arXiv preprint arXiv:1802.01436},
  year={2018}
}

@article{theis2017lossy,
  title={Lossy image compression with compressive autoencoders},
  author={Theis, Lucas and Shi, Wenzhe and Cunningham, Andrew and Husz{\'a}r, Ferenc},
  journal={arXiv preprint arXiv:1703.00395},
  year={2017}
}

@inproceedings{agustsson2019generative,
  title={Generative adversarial networks for extreme learned image compression},
  author={Agustsson, Eirikur and Tschannen, Michael and Mentzer, Fabian and Timofte, Radu and Gool, Luc Van},
  booktitle={Proceedings of the IEEE/CVF international conference on computer vision},
  pages={221--231},
  year={2019}
}

@article{agustsson2020universally,
  title={Universally quantized neural compression},
  author={Agustsson, Eirikur and Theis, Lucas},
  journal={Advances in neural information processing systems},
  volume={33},
  pages={12367--12376},
  year={2020}
}

@inproceedings{galteri2017deep,
  title={Deep generative adversarial compression artifact removal},
  author={Galteri, Leonardo and Seidenari, Lorenzo and Bertini, Marco and Del Bimbo, Alberto},
  booktitle={Proceedings of the IEEE international conference on computer vision},
  pages={4826--4835},
  year={2017}
}

@article{mentzer2020high,
  title={High-fidelity generative image compression},
  author={Mentzer, Fabian and Toderici, George D and Tschannen, Michael and Agustsson, Eirikur},
  journal={Advances in neural information processing systems},
  volume={33},
  pages={11913--11924},
  year={2020}
}

@misc{kodak1999lossless,
  title        = {Kodak Lossless True Color Image Suite},
  author       = {{Eastman Kodak Company}},
  year         = {1999},
  howpublished = {\url{http://r0k.us/graphics/kodak/}},
  note         = {Accessed: 2026-08-04}
}

@inproceedings{agustsson2017ntire,
  title     = {NTIRE 2017 Challenge on Single Image Super-Resolution: Dataset and Study},
  author    = {Agustsson, Eirikur and Timofte, Radu},
  booktitle = {Proceedings of the IEEE Conference on Computer Vision and Pattern Recognition Workshops},
  year      = {2017},
  month     = {July},
  doi       = {10.1109/CVPRW.2017.150}
}

@misc{clic2024,
  title        = {Challenge on Learned Image Compression (CLIC)},
  author       = {{CLIC Organizers}},
  year         = {2024},
  howpublished = {\url{https://clic2024.compression.cc/}},
  note         = {Accessed: 2026-08-04}
}

@inproceedings{wang2003multiscale,
  title     = {Multiscale Structural Similarity for Image Quality Assessment},
  author    = {Wang, Zhou and Simoncelli, Eero P. and Bovik, Alan C.},
  booktitle = {The Thirty-Seventh Asilomar Conference on Signals, Systems \& Computers},
  pages     = {1398--1402},
  year      = {2003},
  doi       = {10.1109/ACSSC.2003.1292216}
}

@inproceedings{zhang2018lpips,
  title     = {The Unreasonable Effectiveness of Deep Features as a Perceptual Metric},
  author    = {Zhang, Richard and Isola, Phillip and Efros, Alexei A. and Shechtman, Eli and Wang, Oliver},
  booktitle = {Proceedings of the IEEE Conference on Computer Vision and Pattern Recognition},
  pages     = {586--595},
  year      = {2018},
  doi       = {10.1109/CVPR.2018.00068}
}

@inproceedings{heusel2017fid,
  title     = {GANs Trained by a Two Time-Scale Update Rule Converge to a Local Nash Equilibrium},
  author    = {Heusel, Martin and Ramsauer, Hubert and Unterthiner, Thomas and Nessler, Bernhard and Hochreiter, Sepp},
  booktitle = {Advances in Neural Information Processing Systems},
  volume    = {30},
  pages     = {6629--6640},
  year      = {2017}
}

@misc{flux-2-2025,
    author={Black Forest Labs},
    title={{FLUX.2: Frontier Visual Intelligence}},
    year={2025},
    howpublished={\url{https://bfl.ai/blog/flux-2}},
}

@misc{blackforestlabs2026flux3,
  title        = {{FLUX 3}: Real World Models: Towards Multimodal Flow Models as the Backbone of Visual Intelligence},
  author       = {{Black Forest Labs}},
  year         = {2026},
  month        = jul,
  day          = {23},
  howpublished = {\url{https://bfl.ai/blog/flux-3}},
  note         = {Accessed: 2026-08-04}
}
\bibliographystyle{iclr2027_conference}

%%%%%%%%%%%%%%%%%%%%%%%%%%%%%%%%%%%%%%%%%%%%%%%%%%%%%%%%%%%%
\clearpage
\appendix

\section{Related Work}
\label{app:related-work}

\subsection{Probabilistic Models and Compression}
\label{subsec:probabilistic-models-compression}
Probabilistic modeling and compression are linked by likelihood optimization: if a shared model assigns probability $p(\mathbf{x})$ to data $\mathbf{x}$, arithmetic coding can encode it using approximately $-\log_2 p(\mathbf{x})$ bits. Maximum-likelihood training can therefore be viewed as minimizing an expected codelength. Recent work makes this connection especially explicit by treating large language models as general-purpose compressors or as probability models for arithmetic coding, including compression across modalities, LLM-generated text, gradients, and interactive lossy text protocols \citep{deletang2024language,haddadi2026comparing,wang2024language,mao2025lossless,rinberg2026haiku}. Related lossless coding views also appear in local bits-back coding for flow models \citep{ho2019compression}, but our focus is lossy compression with generative decoders.
Lossy compression keeps this coding viewpoint but transmits a lower-rate variable rather than the full signal. In generative lossy compression, the decoder uses a shared probabilistic prior to fill in information. DiffC~\citep{theis2022lossy, liu2025lossy} is a representative work that communicates a noisy diffusion variable through reverse channel coding and reconstructs with a shared diffusion prior. 
%Our work follows this probabilistic compression perspective, but proposes an Flux codec with OT path and studies how the Markov transition and reconstruction method affect coding cost, bit allocation, distortion, and perception.

\subsection{Neural Lossy Image Compression}
Neural lossy image compression is commonly built on transform coding based on VAE~\citep{kingma2013auto}. An analysis transform maps an image to quantized latents, an entropy model codes those latents, and a synthesis transform reconstructs the image~\citep{balle2016end, balle2018variational, theis2017lossy, agustsson2020universally}. Recent systems improve either statistical fidelity or perceptual quality by adding stronger likelihood models, adversarial objectives, or generative priors~\citep{agustsson2019generative, galteri2017deep, mentzer2020high}; examples include MS-ILLM \citep{muckley2023improving} and recent very-low-bitrate neural codecs such as PerCo \citep{careil2024perco}.
Diffusion-based codecs replace or augment the decoder with a diffusion prior. DiffC communicates noisy diffusion states and reconstructs by reverse diffusion \citep{theis2022lossy,liu2025lossy}. DDCM uses selected noise-codebook entries as compressed representations \citep{ohayon2025ddcm}, and DiT-IC adapts a pretrained diffusion transformer into a compact one-step compression decoder \citep{shi2026ditic}. Few-step generative codecs formulate short generative trajectories as zero-shot codecs on small images~\citep{kimishima2026fewstep}. In contrast, we focus on a high-resolution Flux-based Rectified Flow codec whose coding transition is defined on the OT path, and analyze how this choice changes coding cost, bit allocation, and reconstruction behavior relative to DiffC.

\subsection{Reverse Channel Coding}
Reverse channel coding (RCC) communicates a sample $\mathbf{x}\sim q(\mathbf{x})$ using a shared proposal distribution $p(\mathbf{x})$ and common randomness between the sender and receiver~\citep{li2018strong}. In the ideal setting, such a sample can be transmitted using approximately $D_{\rm KL}(q\|p)$ bits~\citep{theis2022algorithms,flamich2022fast}. A practical limitation of RCC is its computational cost, since exact or high-quality channel simulation can require many proposal samples. UQDM addresses this issue by replacing the Gaussian channel with a uniform noise channel and using universal quantization, avoiding explicit RCC while preserving a coding cost of NELBO~\citep{yang2024progressive}. A natural next step is to develop universal quantization-based zero-shot compression methods for Rectified Flow, as well as plug-in mechanisms that make such codecs practical for high-resolution images. We leave these directions to future work.

\section{Coding Cost Decomposition and Bit Allocation}
\label{app:coding-cost-bit-allocation}

%We first prove the variational coding-cost decomposition in Eq.~\ref{eq:coding-cost-mismatch}, then derive the FluxC weighted-velocity form and the information-allocation formula over the path parameter.

\subsection{Variational Coding Cost Decomposition}
\label{app:matched-snr-coding-cost}

\lossyrate*

\begin{proof}
By the I-MMSE identity~\citep{guo2005mutual},
\begin{equation}
\frac{d}{d\gamma}I(\mathbf{X};\mathbf{Y}_\gamma)
=
\frac{1}{2\ln 2}\operatorname{mmse}(\gamma),
\qquad
\operatorname{mmse}(\gamma)=
\mathbb{E}\|\mathbf{X}-\hat{\mathbf{X}}_{\mathrm{MMSE}}(\mathbf{Y}_\gamma,\gamma)\|_2^2.
\end{equation}
Integrating from $0$ to the stopping-time SNR $\gamma_t$ gives
\begin{equation}
I(\mathbf{X};\mathbf{Y}_{\gamma_t})-I(\mathbf{X};\mathbf{Y}_0)
=
\frac{1}{2\ln 2}\int_0^{\gamma_t} \operatorname{mmse}(\gamma)d\gamma.
\end{equation}
Since $\mathbf{Y}_0=\boldsymbol{\epsilon}$ is independent of $\mathbf{X}$, $I(\mathbf{X};\mathbf{Y}_0)=0$, and therefore
\begin{equation}
I(\mathbf{X};\mathbf{Y}_{\gamma_t})
=
\frac{1}{2\ln 2}\int_0^{\gamma_t} \operatorname{mmse}(\gamma)d\gamma.
\label{eq:app-integrated-immse}
\end{equation}

Now consider any denoiser $\hat{\mathbf{X}}_\theta(\mathbf{Y}_\gamma,\gamma)$ and define
\begin{equation}
\mathbf{g}_\theta(\mathbf{Y}_\gamma)
=
\hat{\mathbf{X}}_{\mathrm{MMSE}}(\mathbf{Y}_\gamma,\gamma)-\hat{\mathbf{X}}_\theta(\mathbf{Y}_\gamma,\gamma).
\end{equation}
Then
\begin{equation}
\mathbf{X}-\hat{\mathbf{X}}_\theta
=
\mathbf{X}-\hat{\mathbf{X}}_{\mathrm{MMSE}}(\mathbf{Y}_\gamma,\gamma)
+
\mathbf{g}_\theta(\mathbf{Y}_\gamma).
\end{equation}
Squaring and taking expectation gives
\begin{align}
\mathbb{E}\|\mathbf{X}-\hat{\mathbf{X}}_\theta\|_2^2
&=
\mathbb{E}\|\mathbf{X}-\hat{\mathbf{X}}_{\mathrm{MMSE}}(\mathbf{Y}_\gamma,\gamma)\|_2^2
+
\mathbb{E}\|\mathbf{g}_\theta(\mathbf{Y}_\gamma)\|_2^2
\notag\\
&\quad
+2\mathbb{E}\left\langle
\mathbf{X}-\hat{\mathbf{X}}_{\mathrm{MMSE}}(\mathbf{Y}_\gamma,\gamma),
\mathbf{g}_\theta(\mathbf{Y}_\gamma)
\right\rangle .
\end{align}
Since $\mathbf{g}_\theta(\mathbf{Y}_\gamma)$ is measurable with respect to $\mathbf{Y}_\gamma$,
\begin{align}
&\mathbb{E}\left\langle
\mathbf{X}-\hat{\mathbf{X}}_{\mathrm{MMSE}}(\mathbf{Y}_\gamma,\gamma),
\mathbf{g}_\theta(\mathbf{Y}_\gamma)
\right\rangle
\notag\\
&\quad =
\mathbb{E}\left[
\mathbb{E}\left[
\left\langle
\mathbf{X}-\hat{\mathbf{X}}_{\mathrm{MMSE}}(\mathbf{Y}_\gamma,\gamma),
\mathbf{g}_\theta(\mathbf{Y}_\gamma)
\right\rangle
\middle|\mathbf{Y}_\gamma
\right]
\right]
\notag\\
&\quad =
\mathbb{E}\left[
\left\langle
\mathbb{E}[\mathbf{X}-\hat{\mathbf{X}}_{\mathrm{MMSE}}(\mathbf{Y}_\gamma,\gamma)\mid\mathbf{Y}_\gamma],
\mathbf{g}_\theta(\mathbf{Y}_\gamma)
\right\rangle
\right]
\notag\\
&\quad =
\mathbb{E}\left[
\left\langle
\mathbb{E}[\mathbf{X}\mid\mathbf{Y}_\gamma]-\hat{\mathbf{X}}_{\mathrm{MMSE}}(\mathbf{Y}_\gamma,\gamma),
\mathbf{g}_\theta(\mathbf{Y}_\gamma)
\right\rangle
\right]
=0.
\end{align}
Therefore
\begin{equation}
\mathbb{E}\|\mathbf{X}-\hat{\mathbf{X}}_\theta\|_2^2
=
\operatorname{mmse}(\gamma)
+
\mathbb{E}\|\hat{\mathbf{X}}_\theta(\mathbf{Y}_\gamma,\gamma)-\hat{\mathbf{X}}_{\mathrm{MMSE}}(\mathbf{Y}_\gamma,\gamma)\|_2^2.
\label{eq:app-mismatch-pythagorean}
\end{equation}

Following the SNR coordinate formulation of variational diffusion models~\citep{kingma2021variational}, we use the continuous SNR variational coding cost
\begin{equation}
\mathcal{C}_\theta(t)
=
\frac{1}{2\ln 2}
\int_0^{\gamma_t}
\mathbb{E}\|\mathbf{X}-\hat{\mathbf{X}}_\theta(\mathbf{Y}_\gamma,\gamma)\|_2^2 d\gamma.
\label{eq:app-snr-coordinate-coding-cost}
\end{equation}
Substituting Eq.~\ref{eq:app-mismatch-pythagorean} into Eq.~\ref{eq:app-snr-coordinate-coding-cost} yields
\begin{align}
\mathcal{C}_\theta(t)
&=
\frac{1}{2\ln 2}\int_0^{\gamma_t} \operatorname{mmse}(\gamma)d\gamma
+
\frac{1}{2\ln 2}
\int_0^{\gamma_t}
\mathbb{E}\|\hat{\mathbf{X}}_\theta(\mathbf{Y}_\gamma,\gamma)-\hat{\mathbf{X}}_{\mathrm{MMSE}}(\mathbf{Y}_\gamma,\gamma)\|_2^2d\gamma
\notag\\
&=
I(\mathbf{X};\mathbf{Y}_{\gamma_t})
+
\frac{1}{2\ln 2}
\int_0^{\gamma_t}
\mathbb{E}\|\hat{\mathbf{X}}_\theta(\mathbf{Y}_\gamma,\gamma)-\hat{\mathbf{X}}_{\mathrm{MMSE}}(\mathbf{Y}_\gamma,\gamma)\|_2^2d\gamma,
\end{align}
where the last equality uses Eq.~\ref{eq:app-integrated-immse}. 
%This proves Eq.~\ref{eq:coding-cost-mismatch}. The first term is the ideal information cost of the Gaussian observation channel, and the second term is the excess cost from using an imperfect denoiser instead of the posterior mean. In particular, if $\hat{\mathbf{X}}_\theta(\mathbf{Y}_\gamma,\gamma)=\hat{\mathbf{X}}_{\mathrm{MMSE}}(\mathbf{Y}_\gamma,\gamma)$ for all $\gamma$, then $\mathcal{C}_\theta(t)=I(\mathbf{X};\mathbf{Y}_{\gamma_t})$.
\end{proof}

\subsection{FluxC mismatch as weighted velocity error}
\label{app:fluxc-weighted-velocity-cost}

\fluxcmismatch*

\begin{proof}
For the OT path used in FluxC,
\begin{equation}
\mathbf{Z}_t=(1-t)\mathbf{X}+t\boldsymbol{\epsilon}
=\mathbf{X}+t\dot{\mathbf{Z}}_t,
\qquad
\dot{\mathbf{Z}}_t=\boldsymbol{\epsilon}-\mathbf{X}.
\end{equation}
By the denoising reparameterization in Eq.~\ref{eq:rf-denoising-reparam-eps}, the velocity prediction induces
\begin{equation}
\hat{\mathbf{X}}_\theta(\mathbf{Z}_t,t)
=
\mathbf{Z}_t-t\mathbf{v}_\theta(\mathbf{Z}_t,t).
\end{equation}
Using $\mathbf{X}=\mathbf{Z}_t-t\dot{\mathbf{Z}}_t$, the denoising error becomes
\begin{equation}
\|\mathbf{X}-\hat{\mathbf{X}}_\theta(\mathbf{Z}_t,t)\|_2^2
=t^2\|\mathbf{v}_\theta(\mathbf{Z}_t,t)-\dot{\mathbf{Z}}_t\|_2^2.
\label{eq:app-fluxc-velocity-error}
\end{equation}

For the OT path, the SNR is
\begin{equation}
\gamma_t^{\rm OT}=\frac{(1-t)^2}{t^2}.
\end{equation}
Differentiating gives
\begin{align}
\frac{d\gamma_t^{\rm OT}}{dt}
&=-\frac{2(1-t)}{t^3},
\label{eq:app-ot-derivation}
\end{align}

Since $\gamma_t^{\rm OT}$ is the SNR at the stopping time $t$ and $\gamma_1^{\rm OT}=0$, changing variables from the SNR coordinate $\gamma$ to the path time $t$ in Eq.~\ref{eq:app-snr-coordinate-coding-cost} gives
\begin{align}
\mathcal{C}_\theta(t)
&=\frac{1}{2\ln 2}
\int_t^{1}
\mathbb{E}\|\mathbf{X}-\hat{\mathbf{X}}_\theta(\mathbf{Z}_s,s)\|_2^2
\left|\frac{d\gamma_s^{\rm OT}}{ds}\right|ds
\notag\\
&=\frac{1}{2\ln 2}
\int_t^{1}
s^2\mathbb{E}\|\mathbf{v}_\theta(\mathbf{Z}_s,s)-\dot{\mathbf{Z}}_s\|_2^2
\frac{2(1-s)}{s^3}ds
\notag\\
&=
\frac{1}{\ln 2}
\int_t^{1}
\frac{1-s}{s}
\mathbb{E}\|\mathbf{v}_\theta(\mathbf{Z}_s,s)-\dot{\mathbf{Z}}_s\|_2^2ds.
\end{align}
This proves Eq.~\ref{eq:fluxc-coding-cost}. 
%Thus FluxC evaluates the same pointwise velocity target as Rectified Flow, but with the coding-induced time weight $\frac{1-t}{t}$.
\end{proof}

\subsection{Bit allocation over t}
\label{app:bit-allocation-derivation}

\bitallocation*
\begin{proof}
%The goal is to express the information increment per unit path parameter, which supports the rate--time curves in Section~\ref{subsec:experiments-coding-cost-bit-allocation}.
By the I-MMSE identity~\citep{guo2005mutual} and the chain rule,
\begin{align}
\label{app:eq-path-parameter-bit-allocation}
\left|
\frac{d}{dt}I(\mathbf{X};\mathbf{Y}_{\gamma_t})
\right|
&=
\left|
\frac{d}{d\gamma}I(\mathbf{X};\mathbf{Y}_{\gamma})
\right|_{\gamma=\gamma_t}
\left|\frac{d\gamma_t}{dt}\right|
\notag\\
&=
\frac{1}{2\ln 2}\operatorname{mmse}(\gamma_t)
\left|\frac{d\gamma_t}{dt}\right|.
\end{align}

For FluxC, with Eq.~\ref{eq:app-ot-derivation},

\begin{equation}
\left|
\frac{d}{dt}I(\mathbf{X};\mathbf{Y}_{\gamma_t^{\rm OT}})
\right|
=
\frac{1}{\ln 2}
\operatorname{mmse}(\gamma_t^{\rm OT})
\frac{1-t}{t^3}.
\end{equation}

For VP DiffC, from Eq.~\ref{eq:vp-snr},
\begin{equation}
\left|
\frac{d\gamma_t^{\rm VP}}{dt}
\right|
=
\frac{\beta(t)e^{-B(t)}}{(1-e^{-B(t)})^2}.
\end{equation}
Substituting this derivative into Eq.~\ref{app:eq-path-parameter-bit-allocation} gives
\begin{equation}
\left|
\frac{d}{dt}I(\mathbf{X};\mathbf{Y}_{\gamma_t^{\rm VP}})
\right|
=
\frac{1}{2\ln 2}
\operatorname{mmse}(\gamma_t^{\rm VP})
\frac{\beta(t)e^{-B(t)}}{(1-e^{-B(t)})^2}.
\end{equation}
\end{proof}

%These expressions formalize the bit-allocation interpretation: FluxC and DiffC carry the same information at matched SNR, but the information density over coding steps differs because $\gamma_t^{\rm OT}$ and $\gamma_t^{\rm VP}$ have different derivatives. In VP DiffC, this density is controlled by the VP-SDE coefficient $\beta(t)$; in FluxC, it is fixed by the OT path.

\section{Gaussian-Source Diagnostic and Closed Forms}
\label{app:gaussian-closed-form}
\label{app:rd-analysis}
\label{subsec:gaussian-rdp}

This appendix collects the Gaussian source closed forms used in Section~\ref{app:experiments-gaussian}. 
%Closely related calculations for DiffC-A and DiffC-F were given by \citet{theis2022lossy}; we recall the relevant consequences and derive the corresponding OT-path quantities used to interpret FluxC. The goal is diagnostic: the Gaussian source does not model natural images, but it makes the effect of the reconstruction method on distortion and perception analytically transparent.

\subsection{Setup and Metrics}
\label{app:gaussian-setup}

Let
\begin{equation}
\mathbf{X}\sim\mathcal{N}(\mathbf{0},\Sigma),
\qquad
\Sigma=\operatorname{diag}(\lambda_1,\ldots,\lambda_d),
\qquad
\lambda_i>0,
\end{equation}
and $\boldsymbol{\epsilon}\sim\mathcal{N}(\mathbf{0},I)$ be independent of $\mathbf{X}$. We compare reconstruction methods through the common Gaussian observation channel
\begin{equation}
\mathbf{Y}_\gamma=\boldsymbol{\epsilon}+\sqrt{\gamma}\mathbf{X},
\qquad \sqrt{\gamma}\ge0.
\label{eq:gaussian-observation-channel}
\end{equation}
By the standard mutual information formula for a scalar Gaussian channel~\citep{cover2006elements},
\begin{equation}
R_i(\gamma)=I(X_i;Y_{\gamma,i})=\frac12\log_2(1+\gamma\lambda_i).
\end{equation}
so the total rate $R(\gamma)$ is
\begin{equation}
R(\gamma)=\frac12\sum_{i=1}^d\log_2(1+\gamma\lambda_i).
\label{eq:gaussian-observation-rate}
\end{equation}
For the OT path $\mathbf{Z}_t^{\rm OT}=(1-t)\mathbf{X}+t\boldsymbol{\epsilon}$, setting $\gamma=\left(\frac{1-t}{t}\right)^2$ gives $\mathbf{Z}_t^{\rm OT}=t\mathbf{Y}_\gamma$. Hence $I(\mathbf{X};\mathbf{Z}_t^{\rm OT})=I(\mathbf{X};\mathbf{Y}_\gamma)$, because multiplication by a nonzero scalar does not change mutual information.

For a reconstruction $\widehat{\mathbf{X}}_\gamma$ obtained from $\mathbf{Y}_\gamma$, we define its MSE distortion and perception using the squared 2-Wasserstein distance at SNR $\gamma$ as
\begin{align}
D(\gamma)
&:=\mathbb{E}\!\left[\left\|\mathbf{X}-\widehat{\mathbf{X}}_\gamma\right\|_2^2\right],\\
P(\gamma)
&:=W_2^2\!\left(P_{\mathbf{X}},P_{\widehat{\mathbf{X}}_\gamma}\right).
\label{eq:gaussian-perception-metric}
\end{align}
Method-specific subscripts on $D$, $P$, and $\widehat{\mathbf{X}}$ identify the reconstruction rule used below. This choice follows the rate--distortion--perception formulation used in the main text.

\subsection{Posterior-Mean Reconstruction: FluxC-1}
\label{app:gaussian-fluxc-one-step}

\paragraph{Derivation.}
For the affine OT path and $t\in(0,1]$, assume that the velocity regression objective is minimized exactly. Its population minimizer is the ideal conditional velocity field
\begin{equation}
\mathbf{v}_t^\star(\mathbf{z})
:=\mathbb{E}[\dot{\mathbf{Z}}_t\mid \mathbf{Z}_t=\mathbf{z}]
=\mathbb{E}[\boldsymbol{\epsilon}-\mathbf{X}\mid \mathbf{Z}_t=\mathbf{z}],
\end{equation}
where the second equality follows from $\dot{\mathbf{Z}}_t=\boldsymbol{\epsilon}-\mathbf{X}$. Using
$\boldsymbol{\epsilon}=\{\mathbf{Z}_t-(1-t)\mathbf{X}\}/t$, we obtain
\begin{align}
\mathbf{v}_t^\star(\mathbf{z})
&=\mathbb{E}\left[\frac{\mathbf{z}-(1-t)\mathbf{X}}{t}-\mathbf{X}\,\middle|\, \mathbf{Z}_t=\mathbf{z}\right]
\notag\\
&=\frac{\mathbf{z}-\mathbb{E}[\mathbf{X}\mid \mathbf{Z}_t=\mathbf{z}]}{t}.
\end{align}
A single explicit Euler update from time $t$ to $0$, with the velocity evaluated at $(\mathbf{Z}_t,t)$, defines
\begin{align}
\hat{\mathbf{X}}_{{\rm Flux}\text{-}1}
&:=\mathbf{Z}_t-t\mathbf{v}_t^\star(\mathbf{Z}_t)
=\mathbb{E}[\mathbf{X}\mid \mathbf{Z}_t].
\label{eq:app-flux-one-step-posterior-mean}
\end{align}
Hence, under the exact conditional velocity field, the one-step FluxC reconstruction is exactly the posterior mean associated with the communicated variable $\mathbf{Z}_t$. With a learned field $\mathbf{v}_\theta$, this identity is approximate, and the deviation is determined by the velocity-estimation error. The identity itself does not require a Gaussian source; it follows from the affine OT path and the ideal conditional velocity field.

Assume $\mathbb{E}\|\mathbf{X}\|_2^2<\infty$ and define
\begin{equation}
\bar{\mathbf{X}}_t:=\mathbb{E}[\mathbf{X}\mid\mathbf{Z}_t].
\end{equation}
Let $g_t:\mathbb{R}^d\to\mathbb{R}^d$ be any measurable reconstruction rule satisfying $\mathbb{E}\|g_t(\mathbf{Z}_t)\|_2^2<\infty$. Then
\begin{align}
\mathbb{E}\|\mathbf{X}-g_t(\mathbf{Z}_t)\|_2^2
&=\mathbb{E}\|\mathbf{X}-\bar{\mathbf{X}}_t+\bar{\mathbf{X}}_t-g_t(\mathbf{Z}_t)\|_2^2
\notag\\
&=\mathbb{E}\|\mathbf{X}-\bar{\mathbf{X}}_t\|_2^2
+\mathbb{E}\|\bar{\mathbf{X}}_t-g_t(\mathbf{Z}_t)\|_2^2
\notag\\
&\quad+2\mathbb{E}\!\left[\left\langle
\mathbf{X}-\bar{\mathbf{X}}_t,
\bar{\mathbf{X}}_t-g_t(\mathbf{Z}_t)
\right\rangle\right].
\end{align}
The cross term vanishes as
\begin{align}
&\mathbb{E}\!\left[\left\langle
\mathbf{X}-\bar{\mathbf{X}}_t,
\bar{\mathbf{X}}_t-g_t(\mathbf{Z}_t)
\right\rangle\right]
\notag\\
&\quad=\mathbb{E}\!\left[\left\langle
\mathbb{E}[\mathbf{X}-\bar{\mathbf{X}}_t\mid\mathbf{Z}_t],
\bar{\mathbf{X}}_t-g_t(\mathbf{Z}_t)
\right\rangle\right]
=0.
\end{align}
Therefore,
\begin{equation}
\mathbb{E}\|\mathbf{X}-g_t(\mathbf{Z}_t)\|_2^2
=\mathbb{E}\|\mathbf{X}-\bar{\mathbf{X}}_t\|_2^2
+\mathbb{E}\|\bar{\mathbf{X}}_t-g_t(\mathbf{Z}_t)\|_2^2,
\end{equation}
so the unique minimizer up to almost-sure equality is $g_t(\mathbf{Z}_t)=\bar{\mathbf{X}}_t$.

We now specialize to the Gaussian source to obtain closed-form distortion and perception. Since $\mathbf{Z}_t=t\mathbf{Y}_\gamma$ with $\gamma=((1-t)/t)^2$ and $t>0$, the variables $\mathbf{Z}_t$ and $\mathbf{Y}_\gamma$ generate the same sigma-algebra. Therefore,
\begin{equation}
\mathbb{E}[\mathbf{X}\mid\mathbf{Z}_t]
=\mathbb{E}[\mathbf{X}\mid\mathbf{Y}_\gamma].
\end{equation}
Because $\Sigma$ is diagonal, the coordinates are independent and the posterior factorizes. For coordinate $i$,
\begin{equation}
\operatorname{Cov}(X_i,Y_{\gamma,i})=\sqrt{\gamma}\lambda_i,
\qquad
\operatorname{Var}(Y_{\gamma,i})=1+\gamma\lambda_i.
\end{equation}
Thus, the Gaussian conditioning formula gives
\begin{equation}
\hat X_{{\rm Flux}\text{-}1,i}
=\mathbb{E}[X_i\mid\mathbf{Y}_\gamma]
=\mathbb{E}[X_i\mid Y_{\gamma,i}]
=\frac{\sqrt{\gamma}\lambda_i}{1+\gamma\lambda_i}Y_{\gamma,i}.
\end{equation}
The posterior variance in coordinate $i$ is
\begin{align}
\operatorname{Var}(X_i\mid Y_{\gamma,i})
&=\operatorname{Var}(X_i)-
\frac{\operatorname{Cov}(X_i,Y_{\gamma,i})^2}{\operatorname{Var}(Y_{\gamma,i})}
\notag\\
&=\lambda_i-
\frac{\gamma\lambda_i^2}{1+\gamma\lambda_i}
=\frac{\lambda_i}{1+\gamma\lambda_i}.
\end{align}
Therefore, the MSE distortion of FluxC-1 at SNR $\gamma$ is
\begin{equation}
D_{{\rm Flux}\text{-}1}(\gamma)
=\sum_i\frac{\lambda_i}{1+\gamma\lambda_i}.
\label{eq:ot-one-step-distortion}
\end{equation}
The reconstruction marginal is Gaussian with coordinate variance
\begin{equation}
\operatorname{Var}(\hat X_{{\rm Flux}\text{-}1,i})
=\left(\frac{\sqrt{\gamma}\lambda_i}{1+\gamma\lambda_i}\right)^2
\operatorname{Var}(Y_{\gamma,i})
=\frac{\gamma\lambda_i^2}{1+\gamma\lambda_i}.
\end{equation}
For one-dimensional zero-mean Gaussians,
$W_2^2(\mathcal{N}(0,a),\mathcal{N}(0,b))=(\sqrt{a}-\sqrt{b})^2$.
Since the source and reconstruction marginals are diagonal in the same basis, the total squared Wasserstein distance is
\begin{equation}
P_{{\rm Flux}\text{-}1}(\gamma)
=\sum_i\left(
\sqrt{\lambda_i}-
\frac{\sqrt{\gamma}\lambda_i}{\sqrt{1+\gamma\lambda_i}}
\right)^2.
\label{eq:ot-one-step-perception}
\end{equation}
Thus, under the ideal conditional velocity field, FluxC-1 attains the minimum MSE among all square-integrable reconstructions measurable with respect to the communicated observation. This distortion optimality generally comes at a nonzero perception cost because the posterior-mean marginal need not equal $P_{\mathbf{X}}$.

\subsection{Ancestral Sampling and Probability-Flow ODE}
\label{app:gaussian-diffc-reconstructions}

\paragraph{Derivation.}
For VP DiffC, \citet{theis2022lossy} analyze two idealized reconstruction methods for Gaussian sources: ancestral sampling, denoted DiffC-A, and probability-flow ODE reconstruction, denoted DiffC-F. We recall the consequences needed for comparison with FluxC. Their derivation is given for Gaussian sources but in our anisotropic setting, we diagonalize \(\Sigma\) and apply the same one-dimensional Gaussian calculation independently to each eigen coordinate.

Ancestral sampling can be idealized as drawing a fresh posterior sample,
\begin{equation}
\hat{\mathbf{X}}_{{\rm DiffC}\text{-}A}\sim P(\mathbf{X}\mid \mathbf{Y}_\gamma).
\end{equation}

Conditioned on \(\mathbf{Y}_\gamma=\mathbf{y}\), ancestral sampling draws an independent posterior sample
\(\hat{\mathbf{X}}_{{\rm DiffC}\text{-}A}\sim P(\mathbf{X}\mid \mathbf{Y}_\gamma=\mathbf{y})\). Hence, for each coordinate,
\begin{align}
\mathbb{E}\!\left[
(X_i-\hat X_{{\rm DiffC}\text{-}A,i})^2
\mid Y_i=y_i
\right]
&=
\mathbb{E}\!\left[
\{X_i-\mathbb{E}[X_i\mid Y_i]\}
-
\{\hat X_{{\rm DiffC}\text{-}A,i}-\mathbb{E}[X_i\mid Y_i]\}
\right]^2
\notag\\
&=
\operatorname{Var}(X_i\mid Y_i=y_i)
+
\operatorname{Var}(\hat X_{{\rm DiffC}\text{-}A,i}\mid Y_i=y_i)
\notag\\
&=
2\operatorname{Var}(X_i\mid Y_i=y_i).
\end{align}
Using
\[
\operatorname{Var}(X_i\mid Y_{\gamma,i})=\frac{\lambda_i}{1+\gamma\lambda_i},
\]

Hence the distortion of DiffC-A at SNR $\gamma$ is
\begin{equation}
D_{{\rm DiffC}\text{-}A}(\gamma)
=
2\sum_i\frac{\lambda_i}{1+\gamma\lambda_i}.
\label{eq:vp-stochastic-rdp}
\end{equation}

The marginal of a posterior sample is $P_{\mathbf{X}}$, so
\begin{equation}
P_{{\rm DiffC}\text{-}A}(\gamma)=0.
\end{equation}
% This captures the basic distortion--perception tradeoff: ancestral sampling preserves realism in the idealized model, but has larger distortion than the posterior mean because it adds posterior variability.
% The factor of two in Eq.~\ref{eq:vp-stochastic-rdp} also follows from the same conditional-variance decomposition. Let $\hat{\mathbf{X}}_{\mathrm{MMSE}}(\mathbf{Y}_\gamma,\gamma)=\mathbb{E}[\mathbf{X}\mid\mathbf{Y}_\gamma]$. Conditioned on $\mathbf{Y}_\gamma$,
% \begin{align}
% &\mathbb{E}\!\left[
% \|\mathbf{X}-\hat{\mathbf{X}}_{{\rm DiffC}\text{-}A}\|_2^2
% \mid\mathbf{Y}_\gamma
% \right]
% \notag\\
% &\quad=
% \mathbb{E}\!\left[
% \|\mathbf{X}-\hat{\mathbf{X}}_{\mathrm{MMSE}}(\mathbf{Y}_\gamma,\gamma)
% +\hat{\mathbf{X}}_{\mathrm{MMSE}}(\mathbf{Y}_\gamma,\gamma)-\hat{\mathbf{X}}_{{\rm DiffC}\text{-}A}\|_2^2
% \mid\mathbf{Y}_\gamma
% \right]
% \notag\\
% &\quad=
% \mathbb{E}\!\left[
% \|\mathbf{X}-\hat{\mathbf{X}}_{\mathrm{MMSE}}(\mathbf{Y}_\gamma,\gamma)\|_2^2
% \mid\mathbf{Y}_\gamma
% \right]
% +
% \mathbb{E}\!\left[
% \|\hat{\mathbf{X}}_{{\rm DiffC}\text{-}A}-\hat{\mathbf{X}}_{\mathrm{MMSE}}(\mathbf{Y}_\gamma,\gamma)\|_2^2
% \mid\mathbf{Y}_\gamma
% \right]
% \notag\\
% &\quad=
% 2\mathbb{E}\!\left[
% \|\mathbf{X}-\hat{\mathbf{X}}_{\mathrm{MMSE}}(\mathbf{Y}_\gamma,\gamma)\|_2^2
% \mid\mathbf{Y}_\gamma
% \right].
% \end{align}
% The cross term is zero because both conditional residuals have mean zero given $\mathbf{Y}_\gamma$, and the two variance terms are equal because $\mathbf{X}$ and $\hat{\mathbf{X}}_{{\rm DiffC}\text{-}A}$ are conditionally independent posterior samples.

DiffC-F uses the probability-flow ODE associated with the VP diffusion. For the standard normal case, \citet{theis2022lossy} derive the corresponding flow-based distortion. For the multivariate Gaussian source considered here, we apply the same Gaussian probability-flow calculation coordinate-wise in the eigenbasis of $\Sigma$.

The normalized observation has correlation
\begin{align}
r_i(\gamma)
&=\operatorname{Corr}(X_i,Y_{\gamma,i})
\notag\\
&=
\frac{\operatorname{Cov}(X_i,Y_{\gamma,i})}
{\sqrt{\operatorname{Var}(X_i)\operatorname{Var}(Y_{\gamma,i})}} 
\notag \\
&=
\frac{\sqrt{\gamma}\sqrt{\lambda_i}}{\sqrt{1+\gamma\lambda_i}}.
\end{align}
The ideal probability-flow reconstruction is marginal-preserving, so
$\hat X_{{\rm DiffC}\text{-}F,i}\sim\mathcal{N}(0,\lambda_i)$ and hence
\begin{equation}
P_{{\rm DiffC}\text{-}F}(\gamma)=0.
\end{equation}
Applying the standard-normal flow calculation in this coordinate gives
\begin{equation}
\operatorname{Cov}(X_i,\hat X_{{\rm DiffC}\text{-}F,i})
=
\lambda_i r_i(\gamma)
=
\frac{\sqrt{\gamma}\lambda_i^{3/2}}{\sqrt{1+\gamma\lambda_i}}.
\end{equation}
Therefore
\begin{align}
\mathbb{E}\!
\left[(X_i-\hat X_{{\rm DiffC}\text{-}F,i})^2\right]
&=
\operatorname{Var}(X_i)
+
\operatorname{Var}(\hat X_{{\rm DiffC}\text{-}F,i})
-2\operatorname{Cov}(X_i,\hat X_{{\rm DiffC}\text{-}F,i})
\notag\\
&=
2\lambda_i
-2\frac{\sqrt{\gamma}\lambda_i^{3/2}}{\sqrt{1+\gamma\lambda_i}}.
\end{align}
Summing over coordinates gives
\begin{equation}
D_{{\rm DiffC}\text{-}F}(\gamma)
=
2\sum_i\lambda_i
-2\sum_i\frac{\sqrt{\gamma}\lambda_i^{3/2}}{\sqrt{1+\gamma\lambda_i}},
\qquad
P_{{\rm DiffC}\text{-}F}(\gamma)=0.
\label{eq:vp-ode-rdp}
\end{equation}
We use this formula as the VP-ODE reference curve in the Gaussian-source experiment.

\subsection{OT ODE Reconstruction: FluxC-ODE}
\label{app:gaussian-fluxc-ode}

\paragraph{Derivation.}
We now derive the OT counterpart. An ideal many-step OT ODE decoder transports the noisy marginal back to the data marginal. Hence the perception for FluxC-ODE $P_{{\rm Flux}\text{-}ODE}(\gamma)$ is zero.

For Gaussian sources at the same SNR $\gamma$, the many-step OT ODE induces the same Gaussian deterministic coupling as the VP probability-flow ODE. In the eigenbasis of $\Sigma$, each coordinate evolves independently. Since both the source coordinate $X_i$ and the decoded coordinate have variance $\lambda_i$, the coupling is determined by their correlation. The covariance calculation can be made explicit from the matched observation channel. Since $Y_{\gamma,i}=\epsilon_i+\sqrt{\gamma}X_i$,
\begin{equation}
\operatorname{Corr}(X_i,Y_{\gamma,i})
=
\frac{\operatorname{Cov}(X_i,Y_{\gamma,i})}
{\sqrt{\operatorname{Var}(X_i)\operatorname{Var}(Y_{\gamma,i})}}
=
\frac{\sqrt{\gamma}\sqrt{\lambda_i}}{\sqrt{1+\gamma\lambda_i}}.
\end{equation}
The many-step Gaussian ODE reconstruction is marginal-preserving, so $\operatorname{Var}(\hat X_i^{{\rm Flux}\text{-}ODE})=\lambda_i$; the transported coupling therefore has
\begin{equation}
r_i(\gamma)
=\operatorname{Corr}(X_i,\hat X_i^{{\rm Flux}\text{-}ODE})
=\frac{\sqrt{\gamma}\sqrt{\lambda_i}}{\sqrt{1+\gamma\lambda_i}}.
\end{equation}
Thus
\begin{equation}
\operatorname{Cov}(X_i,
\hat X_i^{{\rm Flux}\text{-}ODE})
=\lambda_i r_i(\gamma)
=\frac{\sqrt{\gamma}\lambda_i^{3/2}}{\sqrt{1+\gamma\lambda_i}}.
\end{equation}
Since both $X_i$ and the many-step ODE reconstruction have variance $\lambda_i$, their coordinate-wise squared error is
\begin{equation}
\mathbb{E}\left[(X_i-\hat X_i^{{\rm Flux}\text{-}ODE})^2\right]
=2\lambda_i-2\lambda_i r_i(\gamma).
\end{equation}
Therefore
\begin{equation}
D_{{\rm Flux}\text{-}ODE}(\gamma)
=
2\sum_i\lambda_i
-2\sum_i\frac{\sqrt{\gamma}\lambda_i^{3/2}}{\sqrt{1+\gamma\lambda_i}},
\qquad
P_{{\rm Flux}\text{-}ODE}(\gamma)=0.
\label{eq:ot-ode-rdp}
\end{equation}
Comparing Eqs.~\ref{eq:vp-ode-rdp} and~\ref{eq:ot-ode-rdp}, VP ODE and OT ODE coincide in this matched-SNR Gaussian setting. 
%This equality should not be read as a general equivalence between VP and OT paths. It says that, for a Gaussian source, scalar affine observations, matched SNR, and ideal exact ODE reconstruction, the R--D--P curves do not distinguish the two paths; the main distinction in this diagnostic is the reconstruction method.

\section{Low Rate Expansion for Reconstruction Methods}
\label{app:low-rate-expansion}
\lowratedistortion*

\begin{proof}
We use the SNR-parameterized Gaussian observation channel
\[
\mathbf{Y}_\gamma=\sqrt{\gamma}\,\mathbf{X}+\boldsymbol{\epsilon},
\qquad \gamma\geq0,
\]
and define the corresponding rate in bits as
$R(\gamma):=I(\mathbf{X};\mathbf{Y}_\gamma)$.
By the I--MMSE identity~\citep{guo2005mutual},
\[
\frac{\mathrm{d}}{\mathrm{d}\gamma}R(\gamma)
=\frac{1}{2\ln 2}\operatorname{mmse}(\gamma),
\qquad
\operatorname{mmse}(\gamma)
:=\mathbb{E}\!\left[
\left\|\mathbf{X}-\mathbb{E}[\mathbf{X}\mid\mathbf{Y}_\gamma]\right\|_2^2
\right].
\]
Since $\mathbf{Y}_0=\boldsymbol{\epsilon}$ is independent of the centered source,
$\operatorname{mmse}(0)=\mathbb{E}[\|\mathbf{X}\|_2^2]$.

As $\gamma\downarrow0$, right-continuity of $\operatorname{mmse}(\gamma)$ at
$\gamma=0$ gives
\[
\operatorname{mmse}(\gamma)
=
\operatorname{mmse}(0)+o(1)
=
\mathbb{E}\!\left[\|\mathbf{X}\|_2^2\right]+o(1).
\]
Integrating the I--MMSE identity from $0$ to $\gamma$ gives
\begin{align}
R(\gamma)
&=
R(0)
+
\frac{1}{2\ln 2}
\int_0^\gamma \operatorname{mmse}(s)\,\mathrm{d}s
\notag\\
&=
\frac{1}{2\ln 2}
\int_0^\gamma
\left(
\mathbb{E}\!\left[\|\mathbf{X}\|_2^2\right]+o(1)
\right)\mathrm{d}s
\notag\\
&=
\frac{\mathbb{E}\!\left[\|\mathbf{X}\|_2^2\right]}
{2\ln 2}\gamma
+o(\gamma),
\qquad \gamma\downarrow0,
\end{align}
where we used $R(0)=0$ because
$\mathbf{Y}_0=\boldsymbol{\epsilon}$ is independent of $\mathbf{X}$.

\paragraph{Deterministic ODE reconstruction.}
For the local analysis, parameterize the OT path from data to noise, with the noise endpoint at $t=1$. Its normalized observation has SNR
\[
\gamma_t=\left(\frac{1-t}{t}\right)^2
=(1-t)^2+o\bigl((1-t)^2\bigr),
\qquad t\uparrow1.
\]
Let $\widehat{\mathbf{X}}_{\mathrm{ODE},t}$ denote the output of the ideal ODE decoder initialized at path time $t$. We assume that the decoder admits a one-sided first-order Taylor expansion in $L^2$ at the noise endpoint:
\[
\widehat{\mathbf{X}}_{\mathrm{ODE},t}
=
\widehat{\mathbf{X}}_{\mathrm{ODE},1}
+(1-t)\,\dot{\widehat{\mathbf{X}}}_{\mathrm{ODE},1}
+o_{L^2}(1-t).
\]
At $t=1$, $\widehat{\mathbf{X}}_{\mathrm{ODE},1}$ is independent of $\mathbf{X}$. 
Under the positive first-order source response condition, define
\[
\kappa_{\mathrm{ODE}}
:=
\mathbb{E}\!\left[
\left\langle
\mathbf{X},
\dot{\widehat{\mathbf{X}}}_{\mathrm{ODE},1}
\right\rangle
\right]
>0.
\]
Substituting the \(L^2\) expansion of the decoder output gives
\begin{align}
\mathbb{E}\!\left[
\left\langle
\mathbf{X},
\widehat{\mathbf{X}}_{\mathrm{ODE},t}
\right\rangle
\right]
&=
\mathbb{E}\!\left[
\left\langle
\mathbf{X},
\widehat{\mathbf{X}}_{\mathrm{ODE},1}
\right\rangle
\right]
+
(1-t)\,
\mathbb{E}\!\left[
\left\langle
\mathbf{X},
\dot{\widehat{\mathbf{X}}}_{\mathrm{ODE},1}
\right\rangle
\right]
+o(1-t)
\notag\\
&=
\kappa_{\mathrm{ODE}}(1-t)+o(1-t),
\qquad t\uparrow1.
\label{eq:app-ode-first-order-coupling}
\end{align}
Here, the first term vanishes because
\(\widehat{\mathbf{X}}_{\mathrm{ODE},1}\) is independent of
\(\mathbf{X}\) and both variables are centered.

The ODE distortion can be expanded as
\begin{align}
D_{\mathrm{ODE}}(t)
&:=
\mathbb{E}\!\left[
\left\|
\mathbf{X}
-
\widehat{\mathbf{X}}_{\mathrm{ODE},t}
\right\|_2^2
\right]
\notag\\
&=
\mathbb{E}\!\left[\|\mathbf{X}\|_2^2\right]
+
\mathbb{E}\!\left[
\left\|
\widehat{\mathbf{X}}_{\mathrm{ODE},t}
\right\|_2^2
\right]
-
2\mathbb{E}\!\left[
\left\langle
\mathbf{X},
\widehat{\mathbf{X}}_{\mathrm{ODE},t}
\right\rangle
\right].
\end{align}
Since the ideal ODE decoder preserves the source marginal,
\(\widehat{\mathbf{X}}_{\mathrm{ODE},t}\sim P_X\), and hence
\[
\mathbb{E}\!\left[
\left\|
\widehat{\mathbf{X}}_{\mathrm{ODE},t}
\right\|_2^2
\right]
=
\mathbb{E}\!\left[\|\mathbf{X}\|_2^2\right].
\]
Therefore,
\begin{align}
D_{\mathrm{ODE}}(t)
&=
2\mathbb{E}\!\left[\|\mathbf{X}\|_2^2\right]
-
2\mathbb{E}\!\left[
\left\langle
\mathbf{X},
\widehat{\mathbf{X}}_{\mathrm{ODE},t}
\right\rangle
\right]
\notag\\
&=
D_0-2\kappa_{\mathrm{ODE}}(1-t)
+
o(1-t),
\qquad t\uparrow1,
\label{eq:d-ode-t-one-expansion}
\end{align}
where
\(D_0=2\mathbb{E}[\|\mathbf{X}\|_2^2]\).

Moreover,
\[
R(t):=R(\gamma_t)
=
\frac{\mathbb{E}[\|\mathbf{X}\|_2^2]}{2\ln 2}(1-t)^2
+o\bigl((1-t)^2\bigr).
\]
Since $\mathbb{E}[\|\mathbf{X}\|_2^2]/(2\ln 2)>0$ and $1-t\geq0$ as $t\uparrow1$, taking square roots and inverting this expansion gives
\begin{align}
1-t
=
\sqrt{\frac{2\ln 2}{\mathbb{E}[\|\mathbf{X}\|_2^2]}}\sqrt{R}
+o(\sqrt{R}).
\label{eq:t-reparametrized-r}   
\end{align}

Substituting Eq.~\ref{eq:t-reparametrized-r} into Eq.~\ref{eq:d-ode-t-one-expansion} yields
\[
D_{\mathrm{ODE}}(R)
=
D_0-c_{\mathrm{ODE}}\sqrt{R}
+o(\sqrt{R}),
\]
where
\[
c_{\mathrm{ODE}}
=
2\kappa_{\mathrm{ODE}}
\sqrt{
\frac{2\ln 2}
{\mathbb{E}[\|\mathbf{X}\|_2^2]}
}
>0.
\]

\paragraph{Ancestral sampling.}
Let $\widehat{\mathbf{X}}_{\mathrm{A},\gamma}$ be an ideal posterior sample. Conditioned on $\mathbf{Y}_\gamma$, it is independent of $\mathbf{X}$ and has the same posterior distribution. Therefore,
\[
D_{\mathrm{A}}(\gamma)
:=
\mathbb{E}\!\left[
\left\|\mathbf{X}-\widehat{\mathbf{X}}_{\mathrm{A},\gamma}\right\|_2^2
\right]
=
2\operatorname{mmse}(\gamma).
\]
Right differentiability at zero gives
\[
\operatorname{mmse}(\gamma)
=
\operatorname{mmse}(0)-k\gamma+o(\gamma),\qquad
k:=-\operatorname{mmse}'(0+)\geq0,
\qquad \gamma\downarrow0.\]
Since $D_0=2\operatorname{mmse}(0)$, it follows that
\begin{equation}
D_{\mathrm{A}}(\gamma)
=
D_0-2k\gamma+o(\gamma),
\qquad \gamma\downarrow0.
\label{eq:ancestral-distortion-snr}
\end{equation}
The rate expansion above can be inverted as
\begin{equation}
\gamma
=
\frac{2\ln 2}{\mathbb{E}[\|\mathbf{X}\|_2^2]}R
+o(R),
\qquad R\downarrow0.
\label{eq:low-rate-snr-inversion}
\end{equation}
Substituting Eq.~\eqref{eq:low-rate-snr-inversion} into Eq.~\eqref{eq:ancestral-distortion-snr} gives
\[
D_{\mathrm{A}}(R)
=
D_0-c_{\mathrm{A}}R+o(R),
\qquad R\downarrow0,
\]
where
\[
c_{\mathrm{A}}
=
\frac{4k\ln 2}{\mathbb{E}[\|\mathbf{X}\|_2^2]}
\geq0.
\]
% Equivalently, along the local OT parameterization,
% $D_{\mathrm{A}}(\gamma_t)=D_0-2k(1-t)^2+o((1-t)^2)$: the ancestral improvement is second order in $1-t$, whereas the ODE improvement is first order in $1-t$.

\paragraph{Gaussian-source verification.}
For $\mathbf{X}\sim\mathcal{N}(\mathbf{0},\Sigma)$ with eigenvalues $\{\lambda_i\}_{i=1}^d$,
\[
R(\gamma)
=
\frac{1}{2}\sum_{i=1}^d\log_2(1+\gamma\lambda_i),
\qquad
\operatorname{mmse}(\gamma)
=
\sum_{i=1}^d\frac{\lambda_i}{1+\gamma\lambda_i}.
\]
Thus $k=\sum_i\lambda_i^2$, which verifies the ancestral expansion in closed form. For the ideal ODE decoder, Eqs.~\ref{eq:vp-ode-rdp} and~\ref{eq:ot-ode-rdp} give
\[
\mathbb{E}\!\left[
\left\langle
\mathbf{X},\widehat{\mathbf{X}}_{\mathrm{ODE},t}
\right\rangle
\right]
=
\sum_{i=1}^d
\frac{\sqrt{\gamma_t}\,\lambda_i^{3/2}}
{\sqrt{1+\gamma_t\lambda_i}}
=
(1-t)\sum_{i=1}^d\lambda_i^{3/2}+o(1-t).
\]
Hence $\kappa_{\mathrm{ODE}}=\sum_i\lambda_i^{3/2}>0$, verifying the first-order ODE response in this controlled setting.
\end{proof}

\section{Additional Experiments}
\label{app:additional-experiments}

\begin{figure}[t]
\centering
\includegraphics[width=0.98\linewidth]{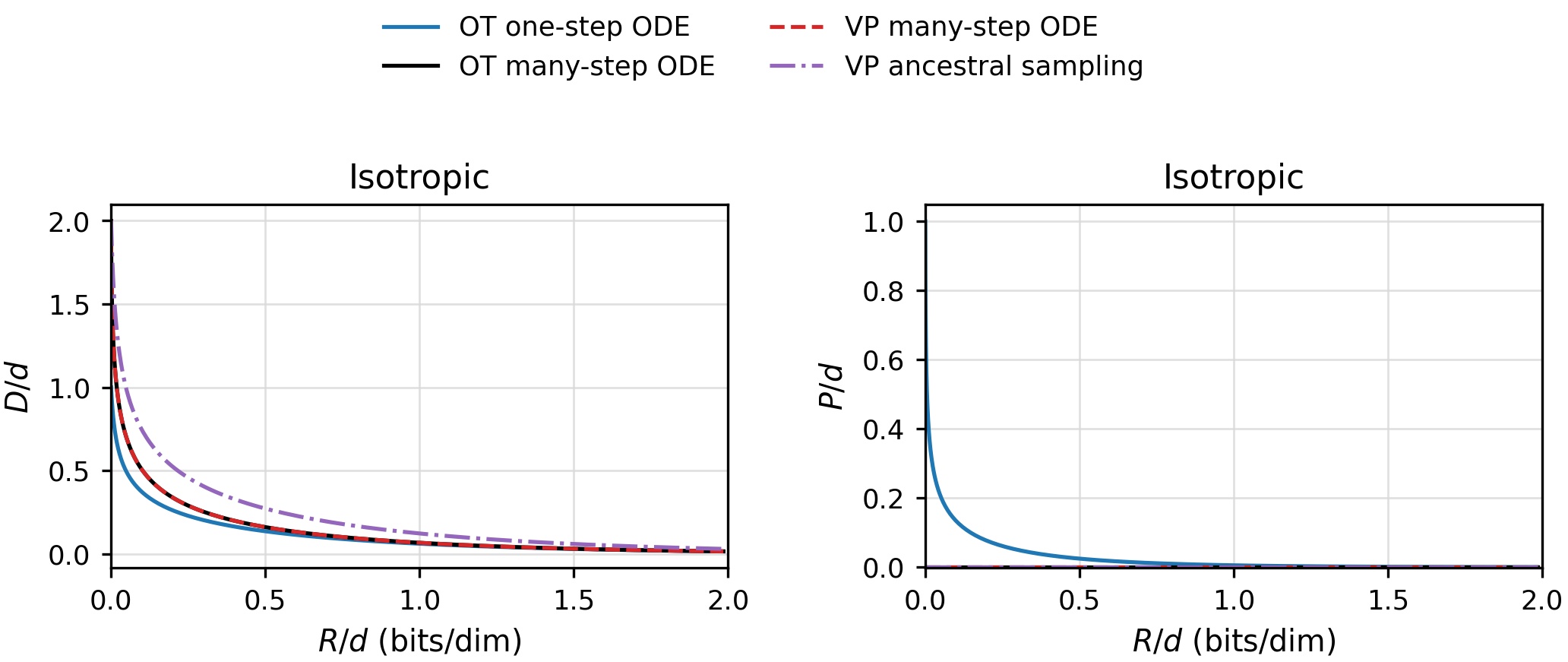}
\caption{Experiments on isotropic Gaussian source. Lower $D/d$ and lower $P/d$ are better.}
\label{fig:gaussian-rdp-spectra}
\end{figure}

\subsection{Gaussian Source}
\label{app:experiments-gaussian}

\paragraph{Setup.}
We use a Gaussian source for the reconstruction analysis in Section~\ref{subsec:reconstruction-methods} and the low rate analysis in Section~\ref{subsec:reconstruction-low-rate}. Although this experiment is not intended to capture the full structure of high-dimensional data, it has an important advantage: the rate--distortion--perception curves can be computed in closed form when the source is Gaussian. This allows us to isolate how the reconstruction method changes distortion and perception at a fixed observation rate, and to test whether the low rate advantage predicted for ODE-based reconstruction appears.
The source is $\mathbf{X}\sim\mathcal{N}(\mathbf{0},\Sigma)$ where \(\Sigma\) is the source covariance matrix. We write its eigenvalues as \(\lambda_1,\ldots,\lambda_d\), so that each \(\lambda_i\) represents the source variance along one principal direction. 
We use four spectra: isotropic with $\lambda_i=1$, mild power law with $\lambda_i\propto i^{-0.5}$, heavy-tailed power law with $\lambda_i\propto i^{-1.25}$, and a low rank spectrum with large variance in the top 32 dimensions and small variance elsewhere. These choices assess whether the reconstruction trends persist as the source changes from uniformly distributed energy to strongly concentrated, low rank structure.
We compare stochastic ancestral sampling, VP ODE, FluxC-1, and FluxC-ODE.
The dimension $d$ is set to 256 and the observation is $\mathbf{Y}_\gamma=\sqrt{\gamma}\,\mathbf{X}+\boldsymbol{\epsilon}$. The metrics are the rate per dimension $R/d$
\begin{equation}
\frac{R(\gamma)}{d}
=
\frac{1}{2d}\sum_{i=1}^d\log_2(1+\gamma\lambda_i),
\label{eq:gaussian-rate-per-dim}
\end{equation}
which is Eq.~\ref{eq:gaussian-observation-rate} normalized by $d$, the distortion per dimension $D/d$, where $D=\mathbb{E}\|\mathbf{X}-\hat{\mathbf{X}}\|_2^2$, and perception per dimension $P/d$ with $P$ defined in Eq.~\ref{eq:gaussian-perception-metric}. Appendix~\ref{app:rd-analysis} derives the closed-form R--D--P curves. For each covariance spectrum, we estimate the distortion and perception curves using \(100{,}000\) independent Monte Carlo samples.

\begin{figure}[t]
\centering
\includegraphics[width=0.98\linewidth]{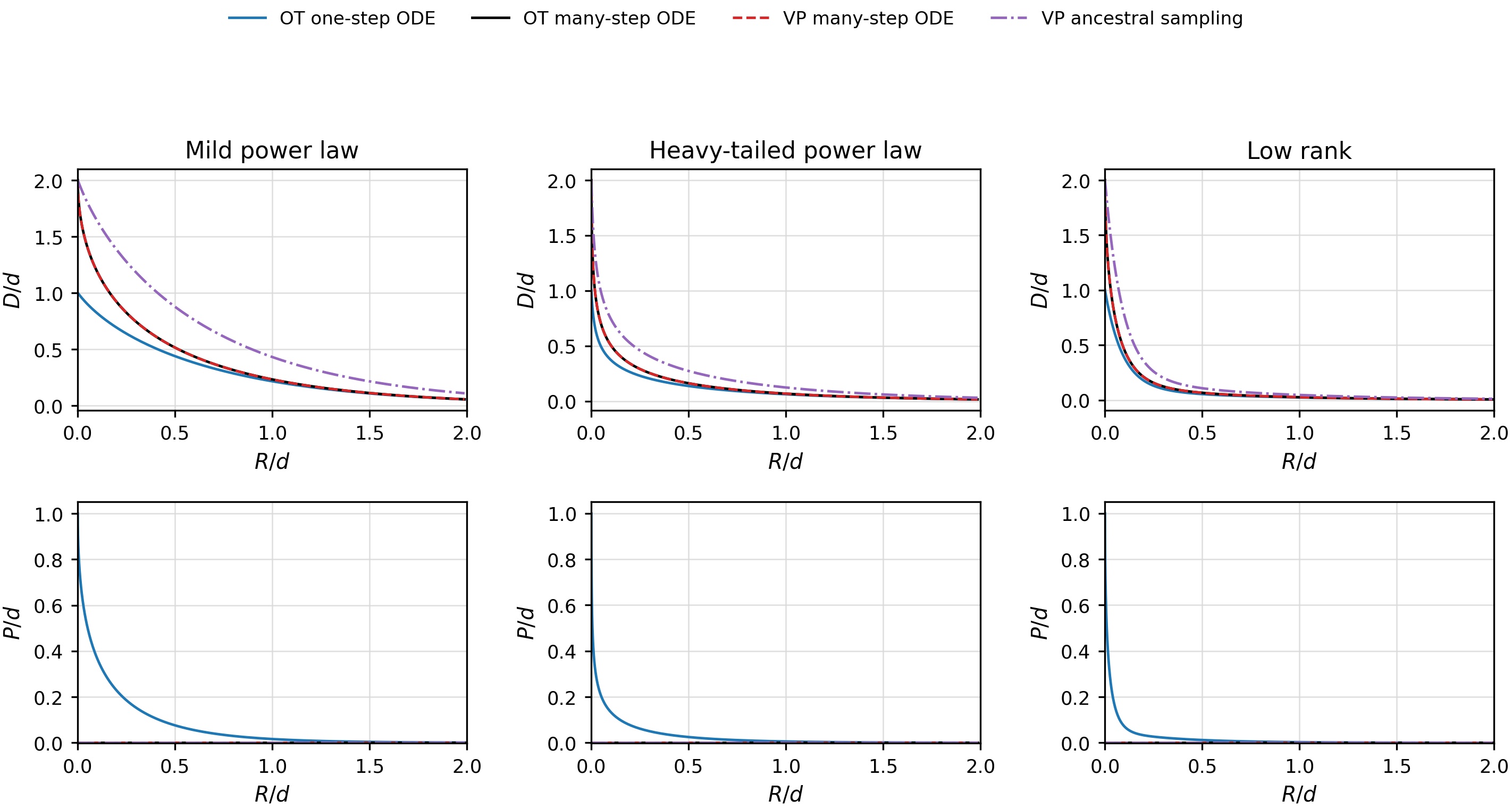}
\caption{Experiments on Gaussian source across mild power-law, heavy-tailed power-law, and low rank spectra.}
\label{fig:gaussian-rdp-spectra-app}
\end{figure}

\paragraph{Results.}
Figure~\ref{fig:gaussian-rdp-spectra} shows three effects for an isotropic Gaussian source: first, VP ODE and FluxC-ODE coincide in this Gaussian setting, as derived in Eqs.~\ref{eq:vp-ode-rdp} and~\ref{eq:ot-ode-rdp}. This confirms that the path selection alone does not explain a distortion gap when the observation SNR and ideal ODE decoder are matched. Second, FluxC-1 attains the lowest distortion at a fixed observation rate, as the one-step OT update coincides with the posterior mean in Eq.~\ref{eq:ot-one-step-distortion}. However, it incurs nonzero perception cost because the reconstruction marginal no longer matches the source marginal. Third, the low rate expansion shows that deterministic ODE reconstruction improves from the zero-rate distortion, whereas stochastic posterior sampling achieves only a slight improvement.
Figure~\ref{fig:gaussian-rdp-spectra-app} shows the results for mild power-law, heavy-tailed power-law, and low rank covariance spectra. Consistent with the isotropic case in Figure~\ref{fig:gaussian-rdp-spectra}, the one-step ODE reconstruction achieves the lowest distortion, with the improvement being most pronounced for the mild power-law spectrum.

\begin{table}[t]
\centering
\caption{Encode and decode time on 512$\times$512 images across three datasets. Times are seconds per image. Encode and decode denote the average processing times per image. Bold indicates the lowest runtime for each dataset.}
\label{tab:encode_decode_time}
\small
\begin{tabular}{llrrrrr}
\toprule
Dataset & Method  & Enc steps & Encode & Decode \\
\midrule
Kodak & DiffC SD1.5  & 50 & 39.41 & 72.82  \\
Kodak & DiffC SD2.1  & 50 & 38.27 & 69.81 \\
Kodak & DiffC Flux  & 46 & 36.00 & 517.08 \\
Kodak & FluxC-1  & 46 & $\mathbf{27.35}$ & $\mathbf{47.91}$ \\
Kodak & FluxC-ODE  & 46 & 35.05 & 574.51 \\
\midrule
DIV2K & DiffC SD1.5  & 50 & 40.10 & 73.14  \\
DIV2K & DiffC SD2.1  & 50 & 38.82 & 70.14 \\
DIV2K & DiffC Flux  & 46 & 36.59 & 517.84\\
DIV2K & FluxC-1  & 46 & $\mathbf{28.24}$ & $\mathbf{51.67}$ \\
DIV2K & FluxC-ODE  & 46 & 40.79 & 555.98 \\
\midrule
CLIC & DiffC SD1.5  & 50 & 39.48 & 73.07  \\
CLIC & DiffC SD2.1  & 50 & 38.61 & 70.44 \\
CLIC & DiffC Flux  & 46 & 35.14 & 517.41 \\
CLIC & FluxC-1  & 46 & $\mathbf{27.20}$ & $\mathbf{47.81}$  \\
CLIC & FluxC-ODE  & 46 & 30.31 & 559.55 \\
\bottomrule
\end{tabular}
\end{table}

\begin{figure}
\centering
\includegraphics[width=0.98\linewidth]{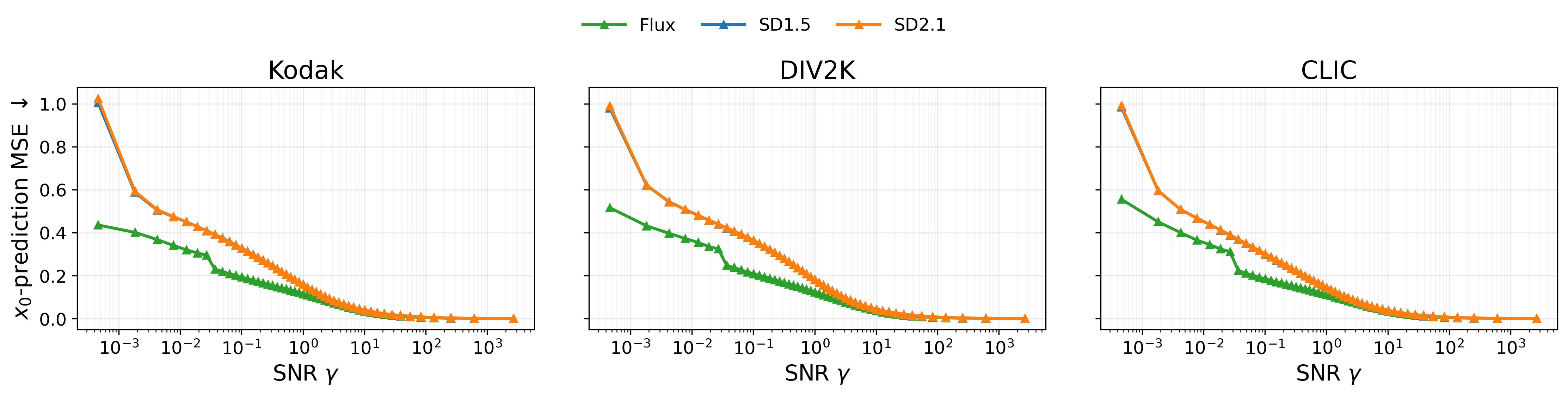}
\caption{Estimation MSE of $\mathbf{x}_0$-prediction as a function of SNR for Flux, SD 1.5, and SD 2.1 on each dataset.}
\label{fig:model-mismatch}
\end{figure}

\begin{table}[t]
\centering
\caption{FluxC encoding and decoding time (s/image) on average. Bold indicates the lowest runtime for each dataset.}
\label{tab:fluxc_step_timing}
\begin{tabular}{llrrr}
\toprule
Dataset & Method & Enc steps & Encode & Decode \\
\midrule
Kodak & FluxC-ODE & 16 & 11.79 & 99.14 \\
Kodak & FluxC-ODE & 6 & 4.57 & 26.59 \\
Kodak & FluxC-ODE & 4 & 5.09 & 9.65 \\
Kodak & FluxC-1 & 16 & 11.67 & 18.48 \\
Kodak & FluxC-1 & 6 & $\mathbf{4.45}$ & 7.69 \\
Kodak & FluxC-1 & 4 & 5.08 & $\mathbf{5.09}$ \\
\midrule
DIV2K & FluxC-ODE & 16 & 12.02 & 98.83 \\
DIV2K & FluxC-ODE & 6 & 4.58 & 26.65 \\
DIV2K & FluxC-ODE & 4 & 5.31 & 9.70 \\
DIV2K & FluxC-1 & 16  & 11.95 & 18.46 \\
DIV2K & FluxC-1 & 6  & $\mathbf{4.56}$ & 7.78 \\
DIV2K & FluxC-1 & 4 & 5.31 & $\mathbf{5.15}$ \\
\midrule
CLIC & FluxC-ODE & 16 & 11.62 & 99.09 \\
CLIC & FluxC-ODE & 6 & 4.46 & 26.72 \\
CLIC & FluxC-ODE & 4 & 5.06 & 9.66 \\
CLIC & FluxC-1 & 16  & 11.59 & 18.40 \\
CLIC & FluxC-1 & 6  & $\mathbf{4.45}$ & 7.74 \\
CLIC & FluxC-1 & 4 & 5.04 & $\mathbf{5.08}$ \\
\bottomrule
\end{tabular}
\end{table}

\begin{figure}[t]
\centering
\includegraphics[width=0.98\linewidth]{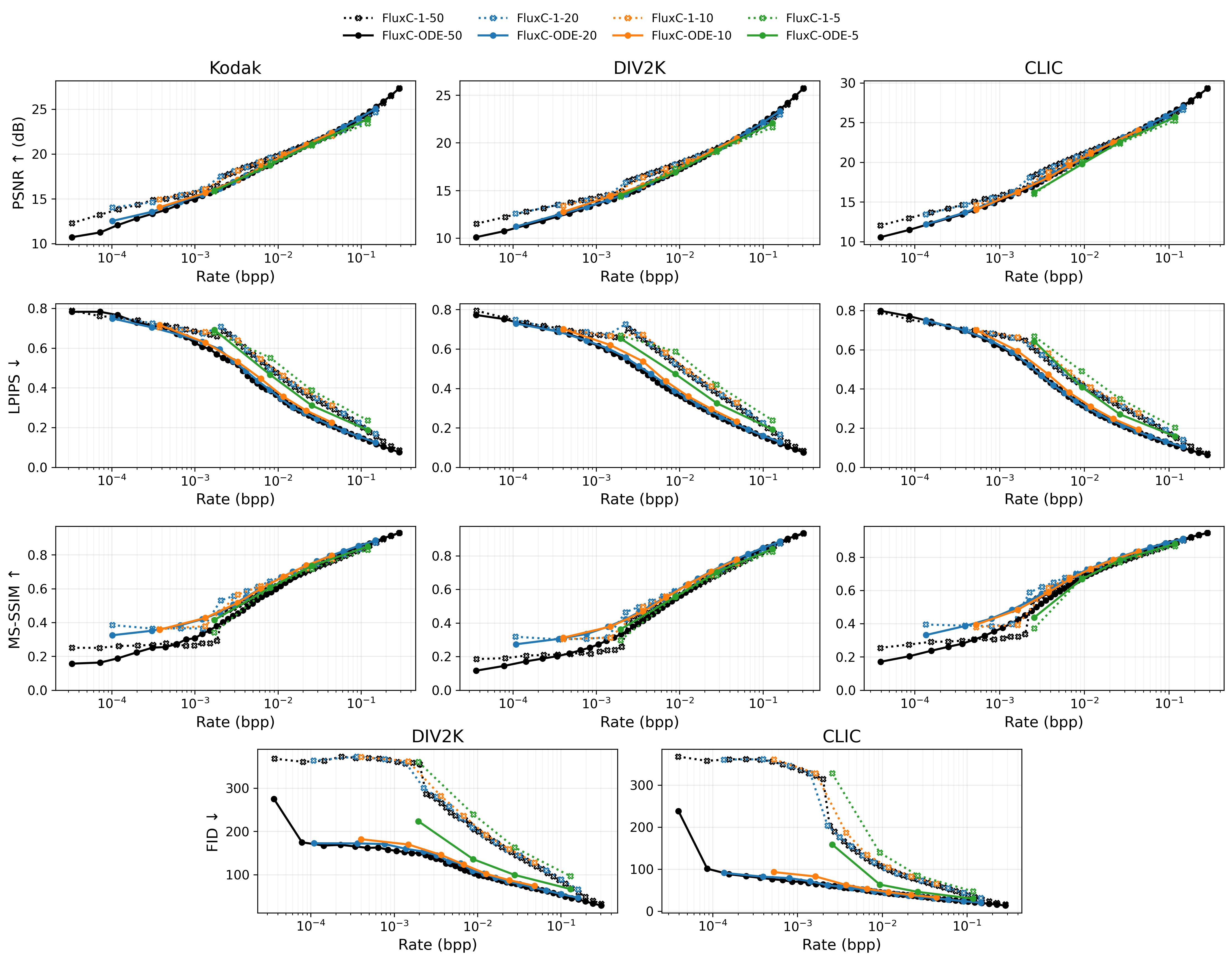}
\caption{Rate--distortion and rate--perception curves for FluxC with nominal encoding schedules of 50, 20, 10, and 5 steps.}
\label{fig:fluxc-step-ablation}
\end{figure}

\subsection{Natural Images}
\label{app:experiments-natural-images}

\paragraph{Computational Efficiency.}
\label{app:computational-efficiency}
Table~\ref{tab:encode_decode_time} reports the average encoding/decoding time per image (seconds) and the number of encoding steps for DiffC and FluxC variants. In FluxC, SNR diverges as the trajectory approaches the data distribution ($t=0$), making communication with RCC impractical~\citep{liu2025lossy}. To address this issue, encoding starts from Gaussian noise at $t=1$ and is terminated after 46 encoding steps, before reaching the data distribution. Consequently, FluxC transmits a lossy latent representation rather than a lossless one. During decoding, the received latent is used to reconstruct the remaining 46 latent samples along the reverse trajectory. For a fair comparison, DiffC Flux also uses 46 encoding steps. By contrast, DiffC employs 50 encoding steps to achieve lossless transmission, while the final latent corresponding to the input data itself is not denoised.

In terms of encoding time, the Markov transition along OT path is slightly faster than the diffusion-based counterpart. For decoding, methods based on Flux require longer runtime because the underlying backbone is substantially larger than SD 1.5 and SD 2.1. Nevertheless, when using a one-step decoder, FluxC can achieve faster reconstruction than DiffC.

\paragraph{Model mismatch.}
\label{app:model-mismatch}
Figure~\ref{fig:model-mismatch} reports the estimation MSE of natural images obtained with SD 1.5, SD 2.1, and Flux across different SNRs for each dataset. SD 1.5 and SD 2.1 exhibit nearly identical errors, while the errors of all three models approach zero as the SNR
increases. In the low SNR regime, $\gamma\leq10^0$, Flux consistently
achieves lower estimation MSE than the SD models. Since the MMSE is
common to all estimators at a fixed SNR, the decomposition in
Observation~\ref{obs:fluxc-velocity-error} attributes this difference to the denoiser-mismatch term. This suggests that FluxC incurs a smaller accumulated mismatch cost, and hence a lower total coding cost at matched low SNR empirically.

\paragraph{Encoding steps.}
\label{app:encoding-steps}
Table~\ref{tab:fluxc_step_timing} reports the computational cost for
nominal encoding schedules of 50, 20, 10, and 5 steps. To avoid the SNR
divergence near the data endpoint, the implementation terminates each
schedule early, resulting in 16, 6, and 4 effective encoding steps,
respectively. Reducing the number of encoding steps generally
decreases both encoding and decoding time, with FluxC-1 providing
particularly fast decoding through its one-step reconstruction.

Figure~\ref{fig:fluxc-step-ablation} presents the corresponding
rate--distortion and rate--perception curves. Using more encoding steps
supports a broader range of rates, particularly in the low rate regime.
At a fixed rate, configurations with more steps consistently achieve
better distortion and perception across the evaluated settings.

% \begin{figure}
% \centering
% \includegraphics[width=0.98\linewidth]{paper_rate_fid_logbpp.jpg}
% \caption{Rate--perception performance comparison.}
% \label{fig:rate-perception-performance-comparison}
% \end{figure}

% \begin{figure}
% \centering
% \includegraphics[width=0.98\linewidth]{iclr27_ultralow_distortion_improvement_all_datasets.jpg}
% \caption{Distortion improvement around low bit rate regime.}
% \label{fig:ultralow-distortion-improvement}
% \end{figure}

\begin{figure}
\centering
\includegraphics[width=0.98\linewidth]{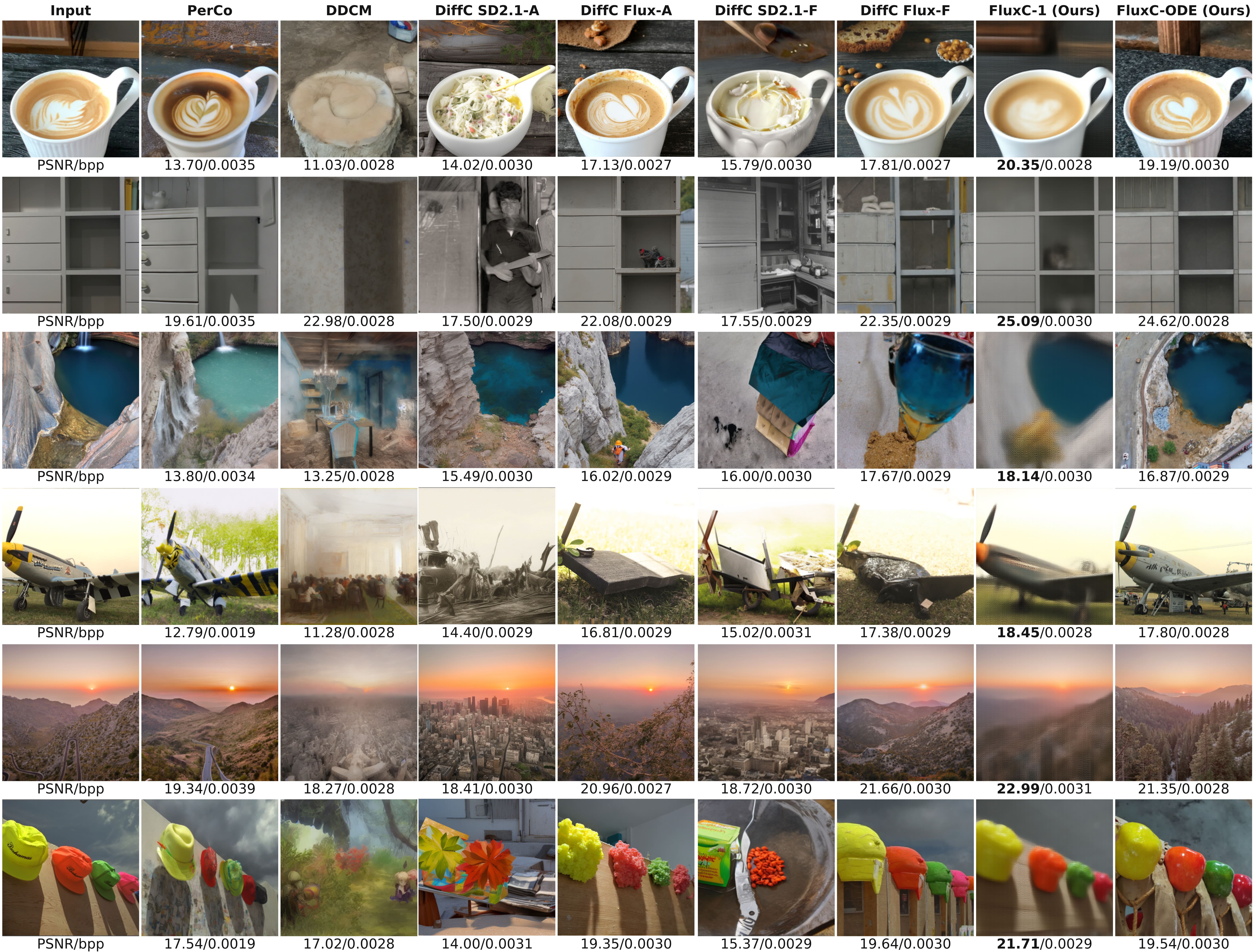}
\caption{Additional qualitative reconstructions at approximately $0.003$ bpp compared with PerCo, DDCM, and DiffC variants and our FluxC variants.}
\label{fig:fluxc-perco-diffc-bpp003-best-cases}
\end{figure}

\begin{figure}
\centering
\includegraphics[width=0.98\linewidth]{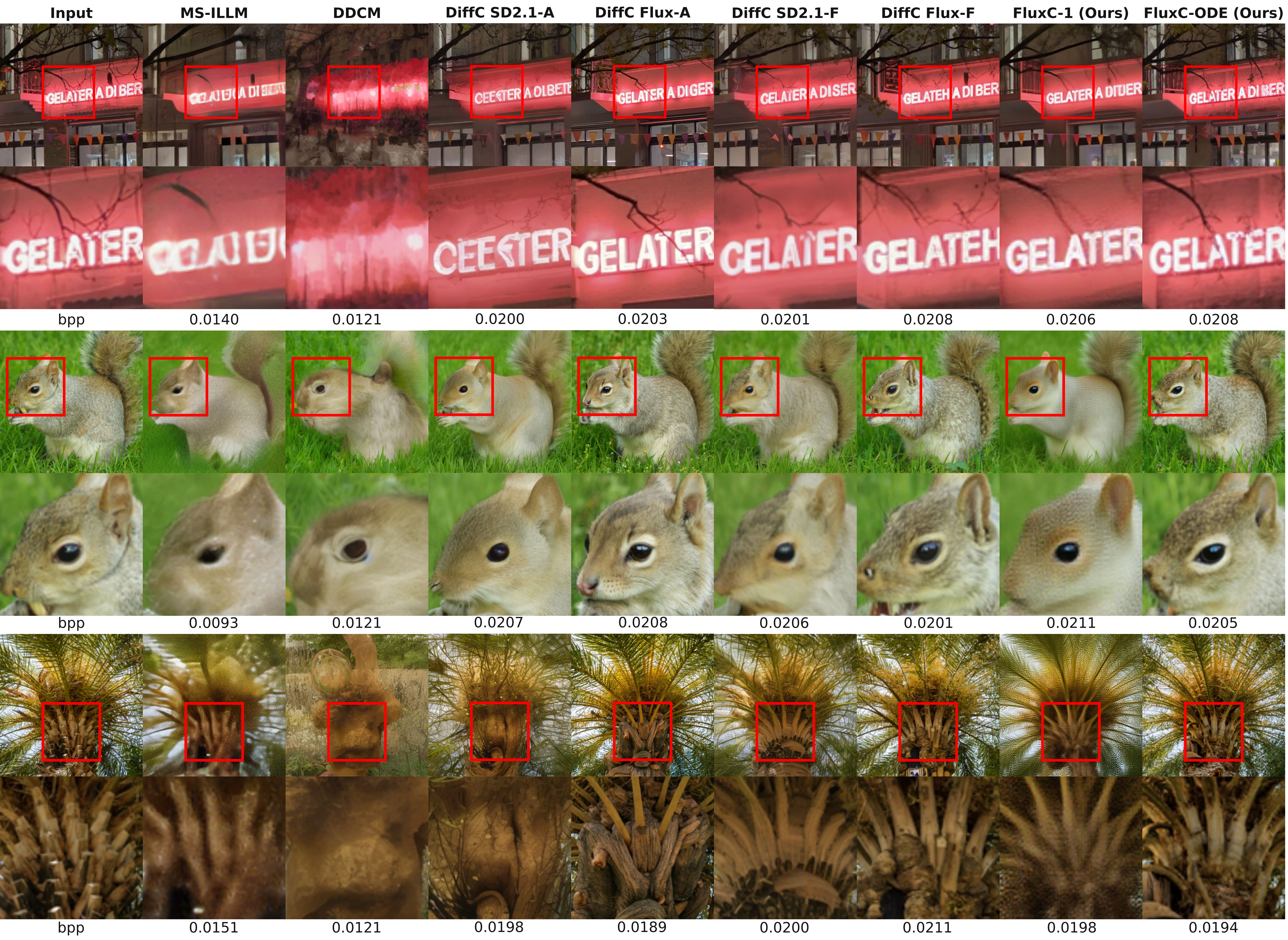}
\caption{Additional qualitative reconstructions at method-specific rates near the \(0.02\) bpp operating region.}
\label{fig:fluxc-baselines-lpips-bpp002-clic-div2k}
\end{figure}

% \paragraph{Rate--perception performance.}
% We show rate--perception performance in Figure~\ref{fig:rate-perception-performance-comparison}. 
% FluxC-1 gives worse FID than the multi-step FluxC decoder, matching the Gaussian-source result in Section~\ref{app:experiments-gaussian}. A posterior-mean reconstruction need not follow the natural-image marginal distribution. It tends to average over plausible images compatible with the transmitted variable, producing smoother and flatter reconstructions. Such reconstructions can reduce pointwise error while moving the reconstruction marginal away from the natural-image distribution, which degrades perception.
% We can see that FluxC-ODE remains competitive with or outperforms existing neural codecs across a broad range of rates.

\paragraph{Additional Qualitative Results.}
\label{app:additional-qualitative-results}
Figure~\ref{fig:fluxc-perco-diffc-bpp003-best-cases} provides additional reconstructions at around $0.003$ bpp for qualitative inspection. 
The overall trend is consistent with that reported in Section~\ref{subsec:experiments-natural}. At these low bit rates, FluxC-1 achieves the lowest distortion, but its reconstructions are typically blurred and resemble averaged images. By contrast, FluxC-ODE may reconstruct fine details that differ slightly from the input, while appearing to better preserve the naturalness of the image.

Figure~\ref{fig:fluxc-baselines-lpips-bpp002-clic-div2k} complements the lower-rate examples by showing reconstructions around $0.02$ bpp. FluxC-ODE better preserves fine details from the input while producing more natural-looking reconstructions, consistent with the lower perceptual distortion (LPIPS) reported in Figure~\ref{fig:natural-rd}.

% \paragraph{Broader impacts.}
% Although FluxC may reduce the storage and communication costs of high-resolution images, its generative decoder may hallucinate or alter visual content, making reconstructed images unsuitable for applications that require faithful preservation of evidence without additional validation.

%%%%%%%%%%%%%%%%%%%%%%%%%%%%%%%%%%%%%%%%%%%%%%%%%%%%%%%%%%%%

\end{document}